\documentclass[conference]{IEEEtran}
\IEEEoverridecommandlockouts
\usepackage[sort]{natbib}
\usepackage{amsmath,amssymb,amsfonts}
\usepackage{algorithmic}
\usepackage{graphicx}
\usepackage{textcomp}
\usepackage{xcolor}
\usepackage{natbib}
\usepackage{hyperref}

\usepackage{latexsym}
\usepackage{enumitem}
\usepackage{algorithm}
\usepackage{color}
\usepackage{amsmath}
\usepackage{amssymb}
\usepackage{mathtools}
\usepackage{amsthm}
\usepackage{wrapfig}
\usepackage{booktabs} % for professional tables
\usepackage{tabularx}
\usepackage{nicefrac}
\usepackage{multirow}
\usepackage{soul}
\usepackage{bm}

\newtheorem{theorem}{Theorem}

\newtheorem{proposition}[theorem]{Proposition}

\newtheorem{definition}{Definition}
\newtheorem{assumption}[theorem]{Assumption}

\def\BibTeX{{\rm B\kern-.05em{\sc i\kern-.025em b}\kern-.08em
    T\kern-.1667em\lower.7ex\hbox{E}\kern-.125emX}}
\begin{document}

\title{FairDP: Achieving Fairness Certification with Differential Privacy
}

\author{
% Anonymous Author(s)
\IEEEauthorblockN{Khang Tran}
\IEEEauthorblockA{
% \textit{Ying Wu College of Computing} \\
\textit{New Jersey Institute of Technology}\\
Newark, New Jersey, USA \\
kt36@njit.edu}
\and
\IEEEauthorblockN{Ferdinando Fioretto}
\IEEEauthorblockA{
% \textit{dept. name of organization (of Aff.)} \\
\textit{University of Virginia} \\
Charlottesville, Virginia, USA \\
fioretto@virginia.edu}
\and
\IEEEauthorblockN{Issa Khalil}
\IEEEauthorblockA{
% \textit{dept. name of organization (of Aff.)} \\
\textit{Qatar Computing Research Institute (QCRI)}\\
Doha, Qatar \\
ikhalil@hbku.edu.qa}
\and
\IEEEauthorblockN{My T. Thai}
\IEEEauthorblockA{
% \textit{dept. name of organization (of Aff.)} \\
\textit{University of Florida}\\
Gainesville, Florida, USA \\
mythai@cise.ufl.edu}
\and
\IEEEauthorblockN{Linh Thi Xuan Phan}
\IEEEauthorblockA{
% \textit{dept. name of organization (of Aff.)} \\
\textit{University of Pennsylvania} \\
Philadelphia, Pennsylvania, USA \\
linhphan@cis.upenn.edu}
\and
\IEEEauthorblockN{NhatHai Phan\textsuperscript{*} \thanks{\textsuperscript{*} Corresponding Author}}
\IEEEauthorblockA{
% \textit{dept. name of organization (of Aff.)} \\
\textit{New Jersey Institute of Technology}\\
Newark, New Jersey, USA \\
phan@njit.edu}
}

\maketitle

\begin{abstract}
    This paper introduces   \textbf{\textsc{FairDP}}, a novel training mechanism designed to provide group fairness certification for the trained model's decisions, along with a differential privacy (DP) guarantee to protect training data. The key idea of \textsc{FairDP} is to train models for distinct individual groups independently, add noise to each group's gradient for data privacy protection, and progressively integrate knowledge from group models to formulate a comprehensive model that balances privacy, utility, and fairness in downstream tasks. By doing so, \textsc{FairDP} ensures equal contribution from each group while gaining control over the amount of DP-preserving noise added to each group's contribution. To provide fairness certification, \textsc{FairDP} leverages the DP-preserving noise to statistically quantify and bound fairness metrics.
    An extensive theoretical and empirical analysis using benchmark datasets validates the efficacy of \textsc{FairDP} and improved trade-offs between model utility, privacy, and fairness compared with existing methods.
    Our empirical results indicate that \textsc{FairDP} can improve fairness metrics by more than $65\%$ on average while attaining marginal utility drop (less than $4\%$ on average) under a rigorous DP-preservation across benchmark datasets compared with existing baselines.
\end{abstract}


\section{Introduction}

Machine learning (ML) systems are being increasingly adopted in decision processes that have a significant impact on people's lives, such as in healthcare, finance, and criminal justice 
\citep{angwin2022machine, giovanola2023beyond}. This adoption also sparked concerns regarding how much information these systems disclose about individuals' data and how they handle bias and discrimination~\citep{fairsurvey, pagano2023bias, wan2023processing}.
% in decision-making processes has brought important considerations regarding privacy, bias, and discrimination. These requirements are becoming pressing as ML systems are increasingly used to make decisions that significantly impact individuals' lives, such as in healthcare, finance, and criminal justice. These concerns underscore the need for ML algorithms that can guarantee both privacy and fairness.

Differential privacy (DP) is an algorithmic property that allows the assessment and bounding of the leakage of sensitive individuals' information during computations. In the context of ML, it enables algorithms to learn from data while ensuring they do not retain sensitive information about any specific individual in the training data. However, directly applying a DP mechanism without careful calibration may aggravate the bias of the trained model's decision to a specific group of data compared to the non-DP ones, which results in unfairness for  different groups of individuals 
\citep{NEURIPS2019_eugene,Fioretto:IJCAI22a,xu2020removing} and incurs societal impacts for such individuals, particularly in areas including finance, criminal justice, or job-hiring~\citep{Cuong:IJCAI21}.
%\Linh{Why? How is it different from bias in ML systems without DP?}

%Fortunately, a growing body of work has recognized the need to study privacy and fairness in learning systems as a combined aspect.
Balancing DP and group fairness while maintaining high model utility in ML systems has been the subject of much discussion in recent years. \citep{cummings19} showed the existence of a trade-off between DP and equal opportunity, a fairness criterion that requires a classifier to have equal true positive rates for different groups. Different studies also reported that when models are trained on data with long-tailed distributions, it is challenging to develop a private learning algorithm that has high accuracy for minority groups \citep{pmlr-v180-sanyal22a}. These findings have led to the question of whether fair models can be created while preserving sensitive information and have spurred the development of various approaches \citep{jagielski:18,mozannar2020fair,Fioretto:NeurIPS21a,tran2020differentially,Fioretto:ArXiv22d}.

While these works have contributed to a deeper understanding of the trade-offs between DP, group fairness, and model utility, as well as the importance of addressing these issues in a unified manner, they all share a common limitation: \emph{the inability to provide formal guarantees for DP and group fairness simultaneously while maintaining high model's utility}. The lack of a formal guarantee for these two critical aspects is essential and cannot be overstated. 
In many critical application contexts, such as those regulated by policy and laws \citep{act2009fair, GDPR, pardau2018california}, these guarantees are often required, and failure to provide them can prevent adoption or deployment. For instance, the Fair Credit Reporting Act \citep{act2009fair} is a federal law that enforces to ensure the fairness and privacy of the information in consumer credit bureau files, which raises a concern in the finance industry around deploying and maintaining more advanced models into production \citep{das2021fairness}. Conversely, a loose theoretical guarantee for fairness and privacy produces a random guess model, which is useless in practice. %\Linh{Give concrete example applications that readers can relate to. The paragraph is an clear -- not sure if ``guarantees'' or "balanced trade-off" is the critical requirement here. Need to better link the two aspects.}

This paper aims to address this gap by proposing a novel training mechanism that significantly improves group fairness with certificates while preserving DP without significantly degrading model utility. 
%\Linh{Utility is abruptly introduced here.} 
The key challenges in developing such a mechanism are: \textbf{(1)} Designing appropriate DP algorithms that can limit the impact of privacy-preserving noise on the model bias; and \textbf{(2)} Balancing the trade-offs between model utility, privacy, and fairness, while simultaneously providing useful fairness certificates. 
%\Linh{Above, we only talked about privacy and fairness tradeoff; need to include utility too for consistency?}

\textbf{Contributions.} The paper makes two main contributions to address these challenges. 
First, it introduces \textsc{FairDP}, a novel DP training mechanism with certified fairness. \textsc{FairDP} remedies the disparate effects of DP-preserving noise on model fairness through group-wise clipping terms, enabling us to derive and tighten certified fairness bounds under DP protection. Throughout the training process, the mechanism progressively integrates knowledge from each group model, significantly improving the trade-off between model utility, privacy, and fairness with upper-bounded utility losses. 
Second, an extensive theoretical and empirical analysis shows that \textsc{FairDP} provides a better balance between privacy and fairness than existing baselines while maintaining high model utility, including both DP-preserving mechanisms with or without fairness constraints.

\section{Background}
We consider datasets $D = \{(x_i, a_i, y_i)\}_{i = 1}^n$ 
whose samples are drawn from an unknown distribution.
Therein, $x_i \in \mathcal{X} \subset \mathbb{R}^d$ is a sensitive feature vector, $a_i \in \mathcal{A} = [K]$ is a (set of) protected group attribute(s), and $y_i \in \mathcal{Y} = \{0,1\}$ is a binary class label, similar to previous work \citep{celis2021fair,icml2022}.
For example, consider a classifier for predicting whether individuals may qualify for a loan. 
The data features $x_i$ may describe the individuals' education, current job, and zip code. The protected attribute $a_i$ may describe the individual's gender or race, and the label $y_i$ indicates whether the individual would successfully repay a loan or not. 
We also use $D_k = \{(x_i, a_i = k, y_i)\}_{i = 1}^{n_k}$ to denote a non-overlapping partition over dataset $D$ which contains exclusively the individuals belonging to a protected group $k$ and $\cap_k D_k = \emptyset$.  
% To ease exposition, we focus on the case of binary protected groups, although the proposed method and analysis are general (see Appx. \ref{appx:extend}).

To generalize for multiple protected group attributes, considering the scenario of $\mathcal{K}$ protected attributes, $\mathcal{A} \subset A_1 \times \dots \times A_\mathcal{K}$ and in each $A_i, i\in[\mathcal{K}]$ there are $K_i$ categories. To apply \textsc{FairDP}, users can divide the dataset $D$ into $K = \prod_{i=1}^{\mathcal{K}}K_i$ disjoint datasets categorized by the combination between the protected attributes. In a particular dataset $D_i = \{x_j, \vec{a}_j, y_j\}_{j=1}^{n_i}, i \in [K]$, each data point $(x_j, \vec{a}_j, y_j)$ will have the protected attribute as $\vec{a}_j \in \mathcal{A}$ and $D_i \cap D_j = \varnothing, \forall i, j \in [K]$. For example, consider a dataset $D$ with the protected attributes are gender with two categories (male and female) and race with five categories (Black, White, Asian, Hispanic, and Other); dataset $D$ can be divided into groups with the combined attributes such as Black male, Black female, Hispanic male, Hispanic female, and so on. Then, users can apply \textsc{FairDP} with the new separation of groups.

% Although the \hai{theoretical or empirical?} results in this paper consider only  one protected attribute \hai{or we just say: Although considering only  one protected attribute in this paper}, \st{the} \hai{our} results can be directly generalized to multiple protected attributes (see \autoref{appx:extend}).

We study models $h_\theta: \mathcal{X} \to [0,1]$ parameterized by $\theta \in \mathbb{R}^r$ and the learning task optimizes the empirical loss function
\begin{equation}
    \mathcal{L}(D) = \min_\theta \sum_{(x_i, a_i, y_i) \in D} \ell \left( h_\theta(x_i), y_i \right),
\end{equation}
where $\ell: \mathcal{Y} \times \mathcal{Y} \to \mathbb{R}_+$ is a differentiable loss function. 

We use $h_\theta$ and $h_{\theta_k}$ to denote, respectively, the models minimizing the empirical loss $\mathcal{L}(D)$ over the entire dataset and that minimizing $\mathcal{L}(D_k)$ using data from the corresponding group $k$. Without loss of generality, consider $h_\theta$ as a combination of a feature extractor $v_\phi$ and a scoring function $u_w$, where $\phi$ and $w$ are their corresponding parameters. Specifically, $v_\phi$ takes $x$ as input and outputs an embedding vector $\xi$, i.e., $\xi = v_\phi(x)$. Then, the scoring function $u_w$, which is a linear layer, takes the embedding $\xi$ and outputs the score $z$ for the prediction on $x$, denoted as follows: $z = u_w(\xi) =  \langle w, \xi\rangle$, where $\langle\cdot,\cdot\rangle$ is the inner product between two vectors. In general, we can write the combination between $u_w$ and $v_\phi$ in $h_\theta$ as follows: $h_\theta(x) = u_w(v_\phi(x))$, where $\theta = \{\phi, w\}$. 
% \khang{In addition, let us denote $A(\cdot)$ as a predictor, which takes the score $z$ and returns the prediction $\hat{y}$ for the corresponding data point (e.g. $\hat{y} = A(z) = 1$ if $z > 0.5$).}
% The prediction $\hat{y} = 1$ if $z > 0$; otherwise, the prediction is $\hat{y} = 0$.

This model setting is general for many structures of modern ML models in classification tasks (e.g., Neural Network, CNN, LSTM, Transformer). Finally, let $\phi_k, w_k$ be the weights of the feature extractor and scoring function of the group $k$'s model~($\theta_k = \{\phi_k, w_k\}$).

\begin{figure}[t]
    \centering
    \includegraphics[width=0.45\linewidth]{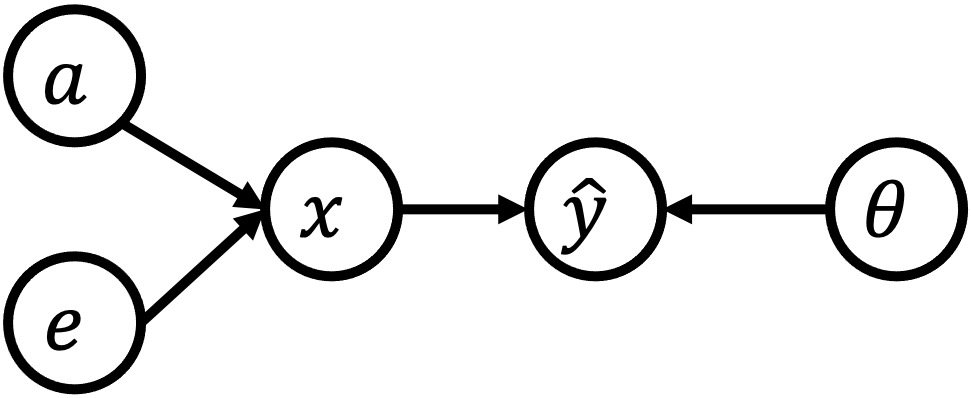}
    \caption{Bayesian network of $h_\theta(x) = u_w(v_\phi(x))$.}
    \label{fig:bayes_net}
\end{figure}

% \begin{wrapfigure}{r}{0.2\textwidth}
% \vspace{-10pt}
%     \begin{center}
%         \includegraphics[width=0.4\columnwidth]{images/bayesian_network.jpg}  % \vspace{-5pt}
%         \caption{Bayesian network of $h_\theta(x) = u_w(v_\phi(x))$.}
%         \label{fig:bayes_net}
%         % \vspace{-20pt}
%     \end{center}
% \end{wrapfigure}

\textbf{Biased Model.} It is worth noting that the model $h_\theta$ only observes the non-protected attributes $x$ as input, which is similar to many practical settings where the protected attribute is hidden due to privacy and/or fairness concerns \citep{chen2019fairness,onesimu2022privacy}. Nevertheless, $x$ is still correlated with the protected attributes $a$, resulting in \textbf{a potentially biased model} affected by the protected attributes $a$ through feature $x$, which leads to unfair predictions as discussed and observed in real-world applications \citep{hajian2012methodology, datta2014automated, corbett2018measure}. Therefore, it is practical to assume that $\hat{y}$ is independent of the protected attribute $a$ and the random event $e$ that depends on group fairness metrics (considered below), given the non-protected attribute $x$. Under this assumption, the predicting process can be mapped by a Bayesian network illustrated in Fig. \ref{fig:bayes_net}

\textbf{Group Fairness and Guarantees.} This paper considers a class of statistical group fairness metrics as follows:

% \nando{I strongly dislike this. We should not call it "general notion of fairness". We should simply say that our methods consider a class of statistical fairness (linear?) metrics.}

\begin{definition}
    The group fairness of a mechanism $\mathcal{M}$ is quantified by
    \begin{equation}
        \mathcal{F} = \max_{u, v \in [K]} [
        Pr(\hat{y} = 1 | a = u, e) - Pr(\hat{y} = 1 | a = v, e) ], \label{eq:fairgeneral} 
    \end{equation} 
    \noindent where $\hat{y}$ is prediction and $e$ is a random event.
\end{definition}
The fairness notion in Eq. \eqref{eq:fairgeneral} captures several well-known group fairness metrics, including demographic parity \citep{fairsurvey} (when $e = \emptyset$), equality of opportunity \citep{hardt:16} (when $e$ is the event ``$y = 1$''), and equality of odd \citep{hardt:16} (when $e = y$). When $\mathcal{F} = 0$, the mechanism $\mathcal{M}$ are said to satisfy \textit{perfect fairness} \citep{williamson2019fairness}. However, perfect fairness cannot be achieved with DP preservation \citep{cummings19}. Therefore, we focus on achieving \textit{approximated fairness}, which allows the fairness metrics to be within a \textit{``meaningful range.''} In addition, if a mechanism satisfies $\mathcal{F} \le \tau$ for $\tau \in [0, 1]$, then we say it \emph{achieves certification of $\tau$-fairness}. Intuitively, as $\tau$ decreases, the model's decision becomes more independent of the protected attribute with respect to different fairness metrics reflected through the random event $e$.

\textbf{Differential Privacy} \citep{dwork2014}. Differential privacy (DP) is a strong privacy concept ensuring that the likelihood of any outcome does not change significantly when a record is added or removed from a dataset. %, thus limiting the amount of information that the outcome reveals about any specific individual.
An adjacent dataset ($D'$) of $D$ is created by adding or removing a record from $D$, denoted as $D \sim D'$.
\begin{definition}[DP]
  \label{dp-def}
  A mechanism $\mathcal{M} \!:\! \mathcal{D} \!\to\! \mathcal{R}$ with domain $\mathcal{D}$ and range $\mathcal{R}$ satisfies $(\epsilon, \delta)$-DP, if, for any two adjacent inputs $D \sim D' \!\in\! \mathcal{D}$, and any subset of outputs $R \subseteq \mathcal{R}$: % \vspace{-2.5pt}
  \[
      \Pr[\mathcal{M}(D) \in R ] \leq  e^{\epsilon} 
      \Pr[\mathcal{M}(D') \in R ] + \delta.
  \] %\vspace{-20pt}
\end{definition}
\noindent 
The parameter $\epsilon > 0$ describes the \emph{privacy loss} of the algorithm, with smaller values denoting stronger privacy, and the parameter 
$\delta \in [0,1)$ is the probability of violating $\epsilon$-DP. 

% The global sensitivity $\Delta_f$ of a real-valued 
% function $f: \mathcal{D} \to \mathbb{R}$  is defined as the maximum amount 
% by which $f$ changes  in two arbitrary adjacent inputs:
% \(
%   \Delta_f = \max_{D \sim D'} \| f(D) - f(D') \|_2.
% \)
% In particular, the Gaussian mechanism, defined by
% \(
%     \mathcal{M}(D) = f(D) + \mathcal{N}(0, \sigma^2\,\mathbf{I}), 
% \)
% \noindent where $\mathcal{N}(0, \sigma^2)$ is 
% the Gaussian distribution with $0$ mean and standard deviation 
% $\sigma^2$, satisfies $(\epsilon, \delta)$-DP for 
% $\sigma =  \Delta_f \nicefrac{\sqrt{2\log(1.25 \// \delta)}}{\epsilon}$. 

\textbf{DPSGD} \citep{abadi:16}. DPSGD is a well-known DP-preserving algorithm to train ML models. The algorithm of DPSGD is illustrated in Algorithm \ref{alg:dpsgd} (Appx. \ref{appx:alg-dpsgd}). At each updating step $t$, DPSGD samples a batch of data using Poisson sampling with a sampling probability $q$. Then, the $l_2$-norm of the gradient derived from each data point in the batch is clipped by a predefined upper-bound $C$ (Line 6).
The \textit{DP-preserving Gaussian} noise with a scale $\sigma$, i.e., $\mathcal{N}(0, C^2\sigma^2\mathbf{I})$, is added to the sum of clipped gradients $\Delta\bar{g}_k$ from all data points, achieving $(q\epsilon, q\delta)$-DP at each step. DPSGD calculates the privacy loss after $T$ steps using a Moment Accountant to track the moment of privacy loss distribution and bound the privacy budget accumulation, given $q$, $\delta$, and $\sigma$.

\section{Related Works}
\label{app:related_works}

Differential privacy has been extensively used in various deep learning applications \citep{phan2016differential,phan2017preserving,phan2017adaptive,userdp,papernot2018scalable,phan2019heterogeneous,phan2020icml}. Meanwhile, numerous efforts have been made to ensure various notions of group fairness through the use of in-processing constraints \citep{feldman2015certifying}, mutual information \citep{gupta2021controllable}, and adversarial training \citep{xu2020theory,jovanovic2022fare,icml2022}. 
A topic of much recent discussion is the implication that DP models may inadvertently introduce or exacerbate biases and unfairness effects on the outputs of a model. 
For example, empirical and theoretical studies have shown that DPSGD can magnify the difference in accuracy observed across various groups, resulting in larger negative impacts for underrepresented groups \citep{NEURIPS2019_eugene,Fioretto:NeurIPS21a, alghamdi2023estimation, islam2023differential}. 
These findings have led to the question of whether it is possible to create fair models while preserving sensitive information. They have spurred the development of various approaches and frameworks such as those presented by \citep{jagielski:18,mozannar2020fair,Fioretto:NeurIPS21a,tran2020differentially, islam2023differential, rafi2024fairness}. 

Despite the advancements made by these efforts, there is limited work addressing the gap in ensuring group fairness. In particular, current methods have not been able to bound the effect of the private models on the model utility in various protected groups. For instance, \citep{mangold2023differential} provides a guarantee that DP mechanisms have bounded disparate impact compared to the non-DP algorithms. However, it is not sufficient to establish a unified understanding of the correlation between differential privacy and group fairness. Similarly, \citep{makhlouf2024impact} studies the impact of local differential privacy on fairness but is unable to establish a unified understanding of the correlation between differential privacy and group fairness. 

To bridge this gap, this paper introduces \textsc{FairDP}, a novel approach to establish a connection between DP preservation and certified group fairness, thereby addressing this crucial challenge. Unlike previous works, \textsc{FairDP} leverages the DP-preserving Gaussian noise added into the gradients of the training process to theoretically provide an upper and a lower bound for the probability that a data point is positively predicted. Then, it introduces a Monte Carlo sampling process under a rigorous DP preservation to approximate the bounds at the inference time, resulting in a tight bound on the group fairness of the model's decisions at the inference time.

\section{Certified Fairness with DP (FairDP)}
\label{sec:fairdp}

\begin{figure}[t]
    \centering
    \includegraphics[width=0.85\linewidth]{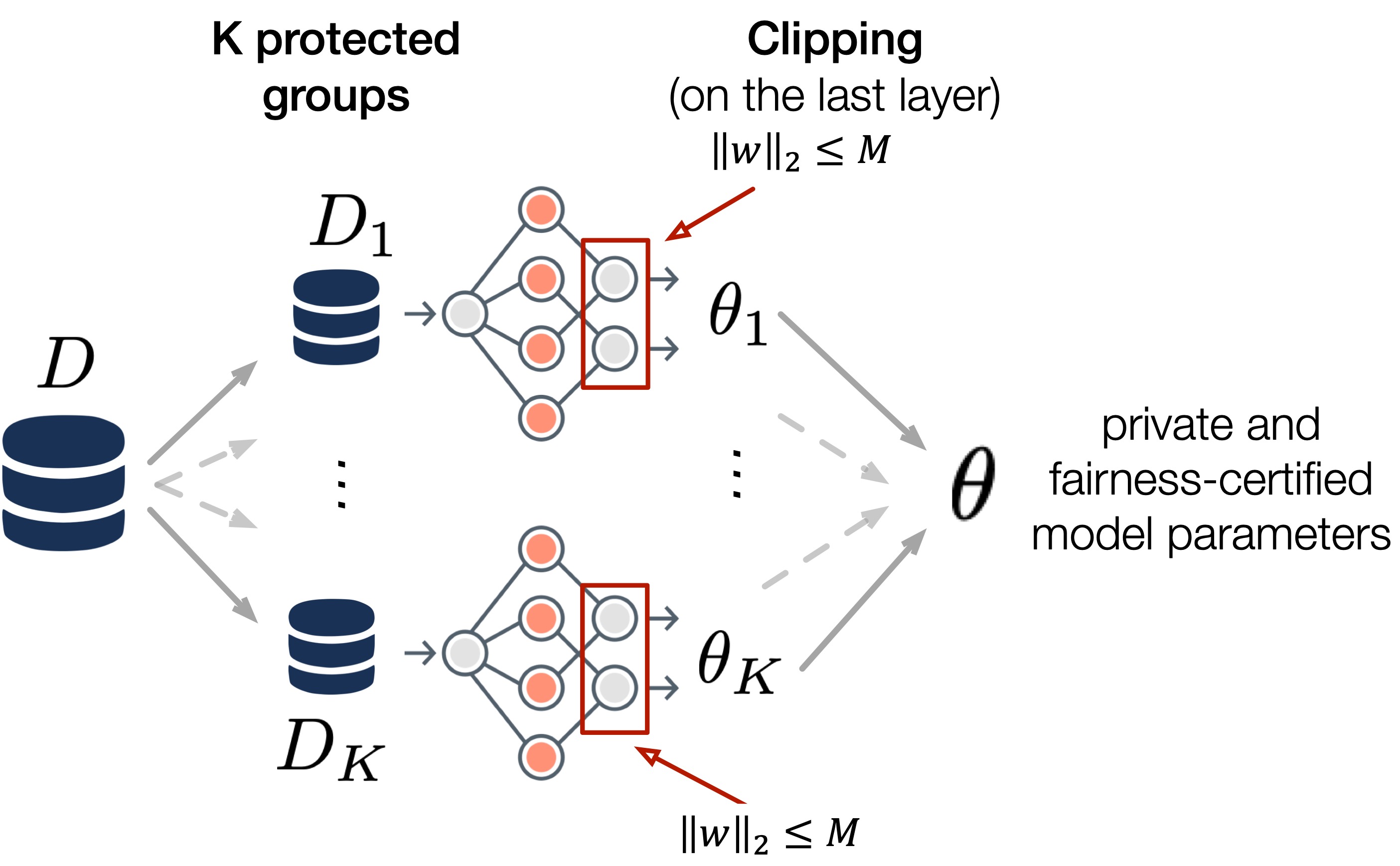}
    \caption{A schematic overview of \textsc{FairDP}.}
    \label{fig:fairDP}
\end{figure}

\begin{algorithm}[t]
    \footnotesize
    \caption{\textsc{FairDP} Training} \label{alg:fairdp}
    \begin{algorithmic}[1]
        \STATE \textbf{Input}: Dataset $D$, sampling rate $q$, noise scale $\sigma$, norm bounds $C$ and $M$, number of steps $T$, learning rate $\eta$, loss function $\ell$.
        % \State \textbf{Output}: Model's parameter $\theta^{(T)}$
        \STATE Initialize $\theta^{0} = \{\phi^0, w^0\}$.
        % \STATE
        % \STATE /* \texttt{Fairness-aware DP Training} */
        \FOR {$t \in [1:T-1]$}
        \STATE \textbf{Clip weights}: $w^{t-1} := w^{t-1}\min(1, M / \|w^{t-1}\|_2)$
        \STATE $\theta^{t-1}_1 = \dots = \theta^{t-1}_K = \theta^{t-1}$ \# model propagation
        \FOR{$k \in \{1, \dots, K\}$}
        \STATE Sample $B^t_k$ from $D_k$ with sampling probability $q$.
        \STATE \textbf{Compute gradient}: For $x_i \in B^t_k$, $g^t_i = \nabla_{\theta^{t-1}_k}\ell(x_i)$
        \STATE \textbf{Clip gradient}: $\bar{g}^t_i = g^t_i\min(1, \frac{C}{\|g^t_i\|_2})$
        \STATE \textbf{Compute total gradient}: $\Delta_k = \sum_{i \in B_k}\bar{g}^t_i$
        \STATE \textbf{Add noise}: $\tilde{\Delta}_k = \Delta_k + \mathcal{N}(0, C^2\sigma^2\mathbf{I})$
        \STATE \textbf{Update}: $\theta^t_k = \theta^{t-1}_k - \frac{\eta}{|B^t_k|}\tilde{\Delta}_k $
        \ENDFOR
        \STATE $\theta^{t} = (\theta^{t}_1 + \dots + \theta^{t}_K)/K$ \# aggregation of groups' models
        \ENDFOR
        \STATE
        \STATE /* \texttt{DP-MC for Fairness Certification} */
        \FOR{$k \in \{1, \dots, K\}$}
            \STATE Sample $B_k$ from $D_k$ with sampling probability $q$.
            \STATE $\{\theta^T_{k,j}\}_{j=1}^N \leftarrow$ \texttt{DP-MC}$(B_k, \eta_T, \sigma, C, M, N)$
        \ENDFOR
        \STATE $\theta^T_{j} = \frac{1}{K} \sum_{k=1}^K\theta^T_{k,j}, \forall j \in [N]$ 
        \STATE \textbf{Return} $\theta^T$
    \end{algorithmic}
\end{algorithm}

This section introduces \textsc{FairDP}, a novel training mechanism satisfying three key objectives: 
{\bf (1)} \emph{Privacy}: the model satisfies $(\epsilon, \delta)$-DP; 
{\bf (2)} \emph{Fairness}: the decisions of released models are unbiased towards any protected group, with theoretical $\tau$-fairness certification; and 
{\bf (3)} \emph{Utility}: the models achieve high utility for downstream ML tasks. 
Achieving these goals is challenging due to the intricate disparate impact incurred by DP-preserving noise and the degradation in the model utility when prioritizing fairness and privacy, particularly without a carefully calibrated noise injection.

% \subsection{FairDP Training}

To overcome these challenges, \textsc{FairDP} relies on two key components, including \textbf{(1)} fairness-aware DP training and \textbf{(2)} Monte Carlo approximation for fairness certification, with a schematic illustration of our training process in Fig. \ref{fig:fairDP} and its pseudo-code in Algorithm \ref{alg:fairdp}. We specifically describe the proposed method and provide the privacy and fairness guarantee in the following sections.

\subsection{Fairness-aware DP Training} 

From the training step $t = 1$ to the step $T-1$ (Lines 4-14), \textsc{FairDP} overcomes a fundamental problem in sampling batches in DPSGD \citep{abadi:16}, which might create a disproportionate number of data points from different groups in the sampled batch, causing disparate contributions from each protected group to the DP-preserving model $h_\theta$. To address this problem, \textsc{FairDP} ensures that every group will contribute to the learning process at any updating steps and upper-bounds the contribution from each group in expectation under privacy protection. This approach remedies the bias of the model's decision toward a group.

To do so, \textsc{FairDP} leverages DPSGD to train a set of group-specific models $\{h_{\theta_k}\}_{k=1}^{K}$, where each $\theta_k$ is independently learned to minimize the loss $\mathcal{L}(D_k)$ of the group $k$ under DP (Lines 8-12). Gradient clipping bounds the $l_2$-norm of the average gradient in each group by $C$ (Line 9); as a result, upper-bounding the contribution from each group in expectation. Also, \textsc{FairDP} clips the $l_2$-norm of the weights of the scoring function $w^{t-1} \in \theta^{t-1}$ by $M$ to narrow the decision boundary in a bounded space (Line 4), which is essential to derive and tighten the fairness certification.

Given the set of group-specific models $\{h_{\theta_k}\}_{k=1}^{K}$ at each training step, \textsc{FairDP} aggregates groups' contributions enabling knowledge distillation from every group to better generalize the model $h_\theta$ (Line 13). Finally, the aggregated model parameters $\theta^{t}$ are propagated as the parameters for every group model in the next training round (Line 5). These aggregation and propagation steps ensure that the general model parameters $\theta^{t}$ are close to the parameters of every group, simultaneously reducing bias towards any specific group and distilling knowledge from every group to improve the model's~utility.

\subsection{DP Monte Carlo \& Ensemble Inference} 

\begin{algorithm}[t]
\footnotesize
\caption{\texttt{DP-MC}}\label{alg:monte_dp}
\begin{algorithmic}[1]
    \STATE \textbf{Input}: Set of clipped gradient $\{g_i | i \in B^T_k\}, \forall k$; learning rate $\eta$; noise scale $\sigma$; norm bounds $C$; number of model $N$; updated $\phi_k^T, \forall k$.
    \FOR{$k \in [K]$}
        \STATE Update $\phi^{T}_{k}$ with the associated DP-preserving gradients.
        \STATE Assemble $\Omega_k = \Big\{\nabla_{w^{T-1}}\ell(x_i)\min\Big(1, \frac{C}{\|g_i\|_2}\Big)\Big\}$ for $x_i \in B^T_k$
        \STATE Partition $\Omega_k$ to $N$ micro-batches $\{G_{k,j}\}_{j=1}^N$ such that $\cup_{j=1}^NG_{k,j} = \Omega_k$ and $G_{k,j}\cap G_{k,j'} = \varnothing, \forall j \neq j'$
        \FOR{$j \in \{1, \dots, N\}$}
            \STATE $\Delta_{k,j} = \sum_{i \in G_j}\nabla_{w^{T-1}}\ell(x_i)\min\Big(1, \frac{C}{\|g_i\|_2}\Big)$
            \STATE $\tilde{\Delta}_{k,j} = \Delta_{k,j} + \mathcal{N}(0, C^2\sigma^2\mathbf{I})$
            \STATE $w^{T}_{k,j} = w^{T-1}_{k,j} - \frac{\eta}{|G_j|}\tilde{\Delta}_{k,j}$
            \STATE $\theta^{T}_{k,j} = [\phi^{T}_{k}, w^{T}_{k,j}]$ 
        \ENDFOR
    \ENDFOR
    \STATE $\theta^T_{j} = \frac{1}{K} \sum_{k=1}^K\theta^T_{k,j}, \forall j \in [N]$ 
    \STATE \textbf{Return} $\{\theta^T_{j}\}_{j=1}^N$
\end{algorithmic} 
\end{algorithm}

% Theorem \ref{theo:worstbound} is anchored on the close-form formulation quantifying the integral $\int_{0}^{+\infty}Pr(z|\xi)dz$. However, it is infeasible to compute this integral for testing data points since the model's parameters are given at inference time. 

% To approximate this integral at inference time

% to sampling the DP-preserving noise added to the gradients of the scoring function to approximate the integral $\int_{0}^{+\infty}Pr(z|\xi)dz$ at the inference time with negligible error while maintaining the same DP protection.

To provide fairness certification at the inference time, we innovate \textsc{FairDP} by executing a DP-preserving Monte Carlo (DP-MC) sampling process at the last step $t = T$ (Lines 16-20) by Algorithm \ref{alg:monte_dp}. The DP-MC samples the DP-preserving noise, which is injected into the gradients of the scoring function to incorporate the randomness of DP-preserving noise into the prediction at inference time. To avoid extra privacy costs and obfuscating the correlation between DP and fairness, i.e., the impact of DP on fairness and vice versa\footnote{Adding multiple noise to the same gradient will incur privacy accumulation \citep{dwork2014}; and adding separate noise to the DP-preserving gradient as smoothing will separate the impact of DP-preserving noise toward the model's decision, obfuscating the correlation between DP and fairness.}, our DP-MC process produces a set of $N$ (general) models $\{h_{\theta^T_j}\}_{j \in [N]}$ by partitioning the batch $B^T_k$ of each group $k$ into $N$ \textit{``disjoint''} micro-batches $\{B^T_{k,j}\}_{j \in [N]}$. For all micro-batches $j \in [N]$, the DP-MC process updates the weights $w_{k,j}$ of the scoring function with DP-preserving gradients $\tilde{\Delta}_{k,j}$ derived from the micro-batch $B^T_{k,j}$. Then, we generate the set of $N$ (general) models $\{h_{\theta^T_j}\}_{j \in [N]}$ by aggregating these weights $\{w_{k,j}\}_{k \in [K], j \in [N]}$ in a group-wise approach, as follows:
\begin{small}\begin{equation}
\forall j \in [N]: w^T_j = \frac{1}{K}\sum_{k = 1}^K w_{k, j}^T \text{\ \ \ and \ \ \ } \theta_j^T = \{\phi^T, w_j^T\}.
\end{equation}
\end{small} 

% Since the micro-batches are disjoint, our DP-MC process does not incur extra privacy.
% \footnote{The parallel composition theorem in DP \citep{parallel}.} 
% compared with Theorem \ref{theo:DP} while sampling the DP-preserving noise $N$ times to approximate the integral $\int_{0}^{+\infty}Pr(z|\xi)dz$. %The pseudo-code of our DP-MC process is in Alg. \ref{alg:monte_dp}, Appx. \ref{appex:alg}.

At the inference time, the prediction on a testing data point $x$ will be the ensemble from the scores of the $N$ models $\{h_{\theta^T_j}\}_{j \in [N]}$. Each model will output a score $z_j$ for $x$, and if the average score is greater than or equal to 0, i.e., $\frac{\sum_{j\in[N]}z_{j}}{N} \ge 0$, then $x$ will be positively predicted, i.e., $\hat{y} = 1$; otherwise,~$\hat{y} = 0$. 

% The ensemble prediction significantly reduces the error in approximating the integral at the rate of $\mathcal{O}(\frac{1}{\sqrt{N}})$ as discussed in \citep{gelman1995bayesian}. Our theoretical analysis is in Appx. \ref{appx:mcerror} and the error is negligible given a small $N = 10$ in our experimental results.

\subsection{DP Guarantee and Fairness Certification} \label{sec:certified} 

We provide the guarantee of DP and $\tau$-fairness certification for the training process of \textsc{FairDP} (Algorithm \ref{alg:fairdp}) in the following theorems.

\textbf{DP Guarantee.} Similar to DPSGD, the DP-guarantee of \textsc{FairDP} is achieved by combining the gradient clipping step and the DP-preserving noise-injecting step. Furthermore, the model propagation and aggregation (Lines 5 and 13) and the weight clipping (Line 4) do not engage with dataset $D$; therefore, they are DP-preserving under the post-processing property of DP \citep{dwork2014}. Furthermore, the DP-MC  separates the gradients at the last epoch to $N$ disjoint sub-batches, which follow the parallel composition theorem \citep{parallel} and do not incur extra privacy risks. Finally, leveraging the Moment Accountant, we can compute the privacy budget $\epsilon$ accumulated throughout $T$ steps.
\begin{theorem}
    Algorithm \ref{alg:fairdp} satisfies $(\epsilon, \delta)$-DP where $\epsilon$ is calculated by the Moment Accountant \citep{abadi:16} given the sampling probability $q$, $T$ steps, and the noise scale $\sigma$.
    \label{theo:DP}
\end{theorem}

\begin{proof}
    Considering one updating step $t$ for an updating process of a particular group $k$, define the gradient extracting function $f(B_{k})$ as follows:
    \begin{align}
        f(B_k) &= \sum_{i \in B_k}g_i\min\Big(1, \frac{C}{\|g_i\|_2}\Big), \text{ if $t\in [1,T-1]$}\\
        f(B_k) &= \Big[\sum_{i \in B_k}\nabla_{\phi}\ell(x_i)\min\Big(1, \frac{C}{\|g_i\|_2}\Big), \nonumber\\
        & \sum_{i \in G_j}\nabla_{w}\ell(x_i)\min\Big(1, \frac{C}{\|g_i\|_2}\Big)\Big]\Bigg\}_{j=1}^{N}, \text{ if $t=T$}
    \end{align}
    where $\{G_j\}_{j=1}^N$ is the disjoint partition of $B_k$.
    
    For the intermediate step $t \in [1,T-1]$, by clipping the gradient, the $l_2$ sensitivity of the total gradient $f(B_k)$ is upper bounded by $C$. 
    Similarly, for the last step $t = T$, for any pairs of neighboring datasets $D$ and $D'$, denote $B_k$ and $ B_k$ as the batches sampled from $D$ and $D'$, respectively. In the worst-case, $B_k$ and $B'_k$ only different at one array of gradient $g_a = [\nabla_{\phi}\ell(x_a), \nabla_{w}\ell(x_a)]$ appears only in one and only one $G_j, j\in [N]$ since $G_i \cap G_i =\varnothing, \forall i,j \in [N]$. Thus, clipping the gradient also ensures the $l_2$ sensitivity of the total gradient $f(B_k)$ is upper-bound by $C$ for the last step. Therefore, we achieve $(q\epsilon, q\delta)$-DP in one updating step by adding Gaussian noise scaled by $C$ to $f(B_k)$ by the argument of Gaussian mechanism \citep{dwork2014}. The model parameter fusing $\theta^{t} = \frac{\theta^{t}_1 + \dots + \theta^{t}_K}{K}$ does not introduce any extra privacy risk at each updating step $t$ following the post-processing property in DP \citep{dwork2014}. We use the moment accountant \citep{abadi:16} to calculate the privacy loss for each dataset $D_k$ after $T$ updating steps given the sampling probability $q$, the broken probability $\delta$, and the noise scale $\sigma$. Finally, since the datasets $\{D_k\}_{k=1}^K$ are disjoint ($D_a \cap D_b = \varnothing, \forall a \neq b \in [1, K]$), by the parallel composition theorem \citep{parallel}, we achieve $(\epsilon, \delta)$-DP for the whole dataset $D$ where $\epsilon$ is calculated by the moment accountant.
\end{proof}

\textbf{Fairness Certification.} To derive a fairness certification, we leverage the DP-preserving noise injected to the parameter of the scoring function $z = u_w(\cdot)$ to bound the probability $Pr(\hat{y}=1|x)$ in a range $[P_{lb}, P_{ub}]$, where $P_{lb}$ and $P_{ub}$ are lower and upper bounds respectively. We focus on the scoring function $u_w$ because it is the decision boundary directly producing the prediction for a data point. Based on that, we can derive $\tau$-fairness certification given the Bayesian network in Fig. \ref{fig:bayes_net}, as follows:
\begin{align}
    \tau &\le P_{ub} - P_{lb}.
\label{tau bound}
\end{align}
The full derivation of Eq. \ref{tau bound} is in the proof of Theorem \ref{theo:worstbound}. 
At step $t$, given a DP-preserving feature extractor $v_{\phi^t}$ yielding the \textit{deterministic} process $\xi = v_{\phi^t}(x)$, the (DP-preserving) Gaussian noise added to the gradients w.r.t. the parameter $w$ transforms the score $z$ into a Gaussian distributed random variable given embedding $\xi$ of an input $x$. Specifically, after injecting Gaussian noise into the gradient (Line 11), the model updating and the aggregating steps (Lines 12 and 13) are linear transformations of the Gaussian noise.
Therefore, $\theta^t$ follows a multivariate Gaussian distribution $\mathcal{N}(w^{t-1} - \eta\mu^t; \sigma_0^2\mathbf{I})$, where $m_k$ is the batch size of $B_k$, $\sigma^2_0 = \frac{\eta^2\sigma^2C^2}{K^2}\sum_{k=1}^K\frac{1}{m^2_k}$, and $\mu^t = \frac{1}{K}\sum_{k=1}^K\mu^t_k$ with $\mu^t_k$ is the total clipped gradient w.r.t. $w^{t-1}$ incurred from $B_k$ of group $k$. 
Also, since $z = \langle w^{t}, \xi\rangle$ is a linear combination of Gaussian random variables of $w^{t}$, the score $z$ becomes a random variable: $z \sim \mathcal{N}(\langle w^{t-1} - \eta\mu^t, \xi\rangle; \|\xi\|_2^2\sigma_0^2)$. 

Note that the weight and gradient clipping \textit{do not affect} the fact that $z$ is Gaussian distributed given $\xi$. It is because \textsc{FairDP} performs the clipping before injecting the Gaussian noise into the gradients. 
As a result, the probability $Pr(\hat{y} = 1 | x) = Pr(z \ge 0 | \xi)$ is an \textit{integral} $\int_{0}^{+\infty}Pr(z|\xi)dz$ over the randomness of the DP-preserving noise added to $w$, which can be computed by a closed-form formula of cumulative distribution function of Gaussian distribution, as follows:
\begin{align}
    Pr(z \ge 0 | \xi) = \frac{1}{2} + \frac{1}{2} \texttt{erf}\Big(\frac{\langle w^{t-1} - \eta\mu^t, \xi\rangle}{\|\xi\|_2\sigma_0\sqrt{2}}\Big), \label{eq:distlarger0} 
\end{align}
where $\texttt{erf}(\cdot)$ is the \textit{error function}. 

By the monotonicity of the error function\footnote{The error function is an increasing function, i.e. if $x_1 < x_2, \forall x_1, x_2 \in \mathbb{R}$, then $\texttt{erf}(x_1) < \texttt{erf}(x_2)$.}, the property of the inner product\footnote{For vector $a, b$, we have $-\|a\|_2\|b\|_2 \le \langle a,b \rangle \le \|a\|_2\|b\|_2$.}, and the fact that the error function is an odd function \footnote{$\forall x \in \mathbb{R}: \texttt{erf}(-x) = -\texttt{erf}(x)$}, we quantify $P_{lb}$ and $P_{ub}$ as follows:
\begin{align}
    P_{lb} = \frac{1}{2} - \frac{1}{2}\texttt{erf}\Big(\frac{\|w^{t-1} - \eta\mu^t\|_2}{\sigma_0\sqrt{2}}\Big), \\
    P_{ub} = \frac{1}{2} + \frac{1}{2}\texttt{erf}\Big(\frac{\|w^{t-1} - \eta\mu^t\|_2}{\sigma_0\sqrt{2}}\Big).
\end{align}

Since the analysis is for a general step $t$, we can derive the fairness certification at the last updating step $T$. However, it is infeasible to compute the integral $\int_{0}^{+\infty}Pr(z|\xi)dz$ over the random variable of DP-preserving noise for testing data points since the model's parameters are given at inference time. Therefore, we innovate the DP-MC to approximate this integral at the inference time and incorporate the randomness of the DP-preserving noise through the ensemble prediction process. Nevertheless, the ensemble prediction significantly incurs the error in approximating the integral at the rate of $\mathcal{O}(\frac{1}{\sqrt{N}})$ as discussed in \citep{gelman1995bayesian}.

Finally, it is worth noting that \textsc{FairDP} clips the weights $w^{t-1}$ and gradients $\mu^t$, i.e., $\|w^{t-1}\|_2 \leq M$ and $\|\mu^t\|_2 \leq C$, we derive $\tau$-fairness certification in the following theorem:
\begin{theorem}
    \label{theo:worstbound}
    \textsc{FairDP} satisfies $\tau$- fairness certification, with
    \begin{equation} 
        \tau \leq {\tt erf}\Big(\frac{MK + \eta C}{K\sigma_0\sqrt{2}}\Big) + \mathcal{O}(N^{-1/2}),
    % \tau \leq {\tt erf}\Big(\frac{(MK + \eta mC)\sqrt{K}}{K\eta\sigma C\sqrt{2}}\Big).
    \label{eq:bound-worst}
    \end{equation}
    where $\sigma_0 = \frac{\eta\sigma C}{K}\sqrt{\sum_{k=1}^K\frac{1}{m^2_k}}$.
\end{theorem}

\begin{proof}
    Recall the considered general fairness metrics:
    \begin{align}
        \mathcal{F} = \max_{u, v \in [K]} [ 
        Pr(\hat{y} = 1 | a = u, e) - Pr(\hat{y} = 1 | a = v, e) ] \nonumber
    \end{align}
    Without loss of generality, assuming that $Pr(\hat{y} = 1 | a = u, e) > Pr(\hat{y} = 1 | a = v, e)$. In the case $Pr(\hat{y} = 1 | a = u, e) < Pr(\hat{y} = 1 | a = v, e)$, we just need to switch the roles of $u$ and $v$. 
    Let $\alpha_k \sim \mathcal{N}(0, \sigma^2C^2\mathbf{I})$ be the DP-preserving noise added to $w_k$ for group $k$. % Furthermore, let $\tilde{\phi}$ be the DP-preserving weight of the feature extractor $v_\phi$ of the general model $h_\theta$. 
    
    The high-level idea to derive a fairness certification is leveraging the DP-preserving noise added to the parameter of the decision-making function $u_w$ to bound the probability $Pr(\hat{y}=1|x)$ by the range $[P_{lb}, P_{ub}]$. Thus, Based on the Bayesian network in Fig. \ref{fig:bayes_net} which , we can derive $\tau$-fairness certification, as follows:
    \begin{align}
        \tau &= \max_{u, v \in [K]} [
        Pr(\hat{y} = 1 | a = u, e) - Pr(\hat{y} = 1 | a = v, e) ] \nonumber\\
        &\le \max_{u, v \in [K]} \Big[
        \int_{x}Pr(\hat{y} = 1 | x)Pr(x|a = u, e)dx \nonumber\\
        &\qquad\qquad\qquad\quad-\int_{x}Pr(\hat{y} = 1 | x)Pr(x|a = v, e)dx \Big] \nonumber\\
        &\le \max_{u, v \in [K]} \Big[
        P_{ub}\int_{x}Pr(x|a = u, e)dx \nonumber \\
        &\qquad\qquad\qquad\qquad\qquad\quad- P_{lb}\int_{x}Pr(x|a = v, e)dx \Big] \nonumber\\
        &= P_{ub} - P_{lb} \nonumber
    \end{align}
    
    To find $P_{ub}$ and $P_{lb}$, we notice that at an updating step $t$, given a DP-preserving updated feature extractor $v_{\phi^t}$ yielding the \textit{\textbf{deterministic}} process $\xi = v_{\phi^t}(x)$, the Gaussian DP-preserving noise added to the gradients w.r.t the parameter $w$ transforms $z$ into a Gaussian distributed random variable given embedding $\xi$ of the input $x$. Let us denote $\mu^t_k$ as the clipped gradient of $w_k$ at the updating step $t$. Indeed, DP-preserving noise injected into clipped gradients $\Delta\bar{g}_k$ transforms the clipped gradients of the scoring function $\mu^t_k$ into a random variable following a multivariate Gaussian distribution $\mathcal{N}(\mu^t_k; \sigma^2C^2\mathbf{I})$. As a result, the parameter of the scoring function of group $k$ at step $t$ becomes a random variable with the following distribution $\mathcal{N}(w^{t-1} - \frac{\eta}{m_k}\mu^t_k; \frac{\eta^2}{m^2_k}\sigma^2C^2\mathbf{I})$ where $m_k = |B_k|$ is the batch size of group $k$, and $w^{t-1}_k = w^{t-1}$.
    
    Furthermore, noticing that $w^t$ of the (general) model is updated by a linear combination of the $K$ multivariate Gaussian random variables $\{w^{t}_k\}_{k\in[K]}$. Therefore, the weight $w^t$ follows a multivariate Gaussian distribution, as follows:
    
    \begin{align}
        &w^t \sim \mathcal{N}\Big(w^{t-1} - \frac{\eta}{K}\sum_{k=1}^K\frac{\mu^t_k}{m_k}; \frac{\eta^2\sigma^2C^2}{K^2}\sum_{k=1}^K\frac{1}{m^2_k}\mathbf{I}\Big). \nonumber
    \end{align}
        
    Denoting $\mu^t = \frac{1}{K}\sum_{k=1}^K\mu^t_k$ and $\sigma^2_0 = \frac{\eta^2\sigma^2C^2}{K^2}\sum_{k=1}^K\frac{1}{m^2_k}$ for simpler notation. Since $z = \langle w^{t}, \xi\rangle$ is a linear combination of the Gaussian random variable, $z$ is a Gaussian distributed random variable, as follows:
    \begin{align}
        z \sim \mathcal{N}\Big(\langle w^{t-1} - \eta\mu^t, \xi\rangle; \|\xi\|_2^2\sigma_0^2\Big).\nonumber
    \end{align}
    % where $\langle\cdot,\cdot\rangle$ is the inner product between two vectors. 
    
    Moreover, under the \textit{\textbf{deterministic}} process $\xi = v_{\phi^t}(x)$, $Pr(\hat{y} = 1 | x) = Pr(\hat{y} = 1 | \xi)$. As a result, given a data point $x$, it will be positively predicted with the probability $Pr(\hat{y} = 1 | x) = Pr(z \ge 0 | \xi) = 1 - Pr(z < 0 | \xi)$, where the probability $Pr(z < 0 | \xi)$ is an \textit{\textbf{integral}} $\int_{-\infty}^{0}Pr(z|\xi)dz$ over the randomness of the DP-preserving noise added to $w$, which a closed-form formula of cumulative distribution function of Gaussian distribution can compute:
    \begin{align}
        Pr(z < 0 | \xi) = \frac{1}{2} + \frac{1}{2} \texttt{erf}\Big(\frac{-\langle w^{t-1} - \eta\mu^t, \xi\rangle}{\|\xi\|_2\sigma_0\sqrt{2}}\Big) \nonumber
    \end{align}
    
    \noindent where $\texttt{erf}(\cdot)$ is the \textit{error function}. It is worth noting that the error function is an odd function \citep{andrews1998special, yang2016engineering}, i.e., $\texttt{erf}(-x) = -\texttt{erf}(x)$ Thus, we have:
    \begin{align}
        Pr(\hat{y} = 1| x) &= Pr(z \ge 0| \xi) = 1 - Pr(z < 0|\xi) \\
        &= \frac{1}{2} + \frac{1}{2} \texttt{erf}\Big(\frac{\langle w^{t-1} - \eta\mu^t, \xi\rangle}{\|\xi\|_2\sigma_0\sqrt{2}}\Big) \nonumber% \label{eq:distlarger0} 
        % &=  \frac{1}{2} + \frac{1}{2}\texttt{erf}\Bigg(\frac{\langle W^{(t-1)} - \eta\mu, z_{L-1}\rangle}{\|W^{(t-1)} - \eta\mu\|_2\|z_{L-1}\|_2} \nonumber\\
        % &\times \frac{\|W^{(t-1)} - \eta\mu\|_2}{\eta\sigma C\sqrt{\frac{2}{K}}}\Bigg), \nonumber 
    \end{align}
    
    Since $\langle w^{t-1} - \eta\mu, \xi\rangle = \|w^{t-1} - \eta\mu\|_2\|\xi\|_2\cos{\varphi}$, with $\varphi$ being the angle between vectors $(w^{t-1} - \eta\mu)$ and $\xi$, $Pr(\hat{y} = 1| x)$ can be computed by:
    \begin{align}
        Pr(\hat{y} = 1| x) &= \frac{1}{2} + \frac{1}{2}\texttt{erf}\Big(\frac{\|w^{t-1} - \eta\mu^t\|_2\cos{\varphi}}{\sigma_0\sqrt{2}}\Big)
        \label{eq:distlarger1-appx}
    \end{align}
    
    From Eq. \eqref{eq:distlarger1-appx}, $\cos(\varphi) \in [-1, 1]$, based on the monotonicity of the error function and the fact that it is an odd function, we have that:

    \begin{align}
        \frac{1}{2} - \frac{1}{2}\texttt{erf}\Big(\frac{\|w^{t-1} - \eta\mu^t\|_2}{\sigma_0\sqrt{2}}\Big)
        \le 
        Pr(\hat{y} = 1| x) \nonumber \\
        Pr(\hat{y} = 1| x) \le \frac{1}{2} + \frac{1}{2}\texttt{erf}\Big(\frac{\|w^{t-1} - \eta\mu^t\|_2\|\xi\|_2}{\|\xi\|_2\sigma_0\sqrt{2}}\Big) \nonumber
    \end{align}

    Therefore, we have that:

    \begin{align}
        P_{lb} = \frac{1}{2} - \frac{1}{2}\texttt{erf}\Big(\frac{\|w^{t-1} - \eta\mu^t\|_2}{\sigma_0\sqrt{2}}\Big) \label{eq:lowerboundl2} \\
        P_{ub} = \frac{1}{2} + \frac{1}{2}\texttt{erf}\Big(\frac{\|w^{t-1} - \eta\mu^t\|_2}{\sigma_0\sqrt{2}}\Big) \label{eq:upperboundl2}
    \end{align}

    Now, we can leverage Eq. \eqref{eq:lowerboundl2} and Eq. \eqref{eq:upperboundl2} to indicate that:
    \begin{align}
        Pr(\hat{y} = 1 | a=k, e) &\le \mathbb{E}_{x|a,e}\Big[\frac{1}{2} + \frac{1}{2}\texttt{erf}\Big(\frac{\|w^{t-1} - \eta\mu^t\|_2}{\sqrt{2}\sigma_0}\Big)\Big] \nonumber\\
        &= \frac{1}{2} + \frac{1}{2}\texttt{erf}\Big(\frac{\|w^{t-1} - \eta\mu^t\|_2}{\sqrt{2}\sigma_0}\Big) \nonumber \\
        Pr(\hat{y} = 1 | a=k, e) &\ge \mathbb{E}_{x|a,e}\Big[\frac{1}{2} - \frac{1}{2}\texttt{erf}\Big(\frac{\|w^{t-1} - \eta\mu^t\|_2}{\sqrt{2}\sigma_0}\Big)\Big] \nonumber\\
        &= \frac{1}{2} - \frac{1}{2}\texttt{erf}\Big(\frac{\|w^{t-1} - \eta\mu^t\|_2}{\sqrt{2}\sigma_0}\Big) \nonumber
    \end{align}
    As a result, we have:
    \begin{align}
        \mathcal{F} &= \max_{u, v \in [K]} [ 
        Pr(\hat{y} = 1 | a = u, e) - Pr(\hat{y} = 1 | a = v, e) ] \nonumber \\
        &\le \texttt{erf}\Bigg(\frac{\|w^{t-1} - \eta\mu^t\|_2}{\sqrt{2}\sigma_0}\Bigg) \nonumber
    \end{align}
    By the monotonicity of the error function, we have
    \begin{align}
        \mathcal{F} &\le \texttt{erf}\Bigg(\frac{\|w^{t-1} - \eta\mu^t\|_2}{\sqrt{2}\sigma_0}\Bigg) \le \texttt{erf}\Bigg(\frac{\|w^{t-1}\|_2 + \eta\|\mu^t\|_2}{\sqrt{2}\sigma_0}\Bigg) \nonumber
    \end{align}
    Furthermore, by the clipping process in Lines 6 and 11 of Algorithm \ref{alg:fairdp}, we have 
    \begin{align}
        &\|w^{t-1}\|_2 \le = M \nonumber \\
        &\|\mu^t\|_2 \le \frac{1}{K}\sum_{k=1}^K\frac{\|\mu^t_k\|_2}{m_k} \le \frac{1}{K}\sum_{k=1}^K\frac{\sum_{i \in B_k}\|\bar{g}_i\|_2}{m_k} \nonumber\\
        &\qquad\le \frac{1}{K}\sum_{k=1}^K\frac{\sum_{i \in B_k}C}{m_k} = \frac{C}{K} \nonumber
    \end{align}
    Therefore, along with the MC approximation error, we have the worst-case fairness certification on the true data distribution of different groups and the DP-preserving noise distribution, as follows:
    \begin{align}
        \mathcal{F} &\le \texttt{erf}\Big(\frac{(MK + \eta C)}{K\sigma_0\sqrt{2}}\Big) + \mathcal{O}(N^{-1/2}) \nonumber
    \end{align}
    which concludes the proof.
\end{proof}

% \noindent\framebox{
% \begin{minipage}{.96\linewidth}
\textit{\textbf{Remark 1.} 
Theorem \ref{theo:worstbound} provides an upper bound on the $\tau$-fairness certification, revealing a novel insight into the trade-off among privacy, fairness, and utility. The upper bound of $\tau$-fairness certification decreases as the DP-preserving noise scale $\sigma$ increases. 
As a result, stronger privacy (larger $\sigma$) enhances fairness certification due to the increased randomness influencing the model's decisions. Our theoretical observation is consistent with previous studies \citep{xu2019achieving,pannekoek2021investigating}.}

\subsection{Tightening Fairness Certification}
\label{Expected Fairness}

While an important result, larger batch sizes, and lower learning rates can result in a looser $\tau$-fairness in Theorem \ref{theo:worstbound}. Furthermore, the bound in Theorem \ref{theo:worstbound} is guaranteed in the worst-case scenario when the model is completely biased to a specific group (i.e., the model consistently predicting positive for a group and negative for other groups), which might not be tight for a given data domain. To overcome this issue, we derive an empirical fairness certification that substantially tightens the $\tau$-fairness certification, enabling a better understanding of the privacy, fairness, and utility trade-offs. Specifically, noticing that the probability that a data point from a group $k$ is positively predicted, conditioned on a random event $e$, can be quantified as follows:
\begin{align}
    Pr(\hat{y} = 1 | k, e) &= \int_{x}Pr(\hat{y}=1|x)Pr(x| k, e)dx. \nonumber
\end{align}
By leveraging Eq. \eqref{eq:distlarger0}, this probability can be quantified by the expectation $\mathbb{E}_{x \sim P(x| k, e)}\Big[\frac{1}{2} + \frac{1}{2} \texttt{erf}\Big(\frac{\langle w^{t-1} - \eta\mu^t, \xi\rangle}{\|\xi\|_2\sigma_0\sqrt{2}}\Big)\Big]$, taken over by the data distribution of group $k$. Assuming that the training data and the testing data at the inference time for any group $k$ are from the same distribution (i.e., marginal or no distribution shift), which is practical by the stability of DP mechanisms \citep{kulynych2022you}, one can approximate this expectation using the training data and provide a bound within a confidence interval, as follows:
\begin{small}
\begin{align}
    &\hat{\mathbb{E}}_{k,e} = \frac{1}{2} + \frac{1}{2n_{k, e}}\sum_{x \in D_{k, e}}\texttt{erf}\Big(\frac{\langle w^{t-1} - \eta\mu, v_{\phi^t}(x)\rangle}{\|v_{\phi^t}(x)\|_2\sigma_0\sqrt{2}}\Big), \nonumber
\end{align}
\end{small}

\noindent where $D_{k,e}$ is the subset of $D_k$ satisfying the random event $e$, and $n_{k,e}$ is the size of $D_{k,e}$. 
For instance, $D_{k,e} = D_{k}$ for \textbf{\textit{demographic parity}}, $D_{k, e}$ is the set of data point in $D_k$ with the positive label for \textbf{\textit{equality of opportunity}}, and $D_{k, e}$ is the set of data point in $D_k$ with the positive label when computing true positive rate or the negative label when computing false positive rate for \textbf{\textit{equality of odd}}. 

Finally, the empirical $\tau$-fairness certification can be computed by $\max_{u, v \in [K]}[\hat{\mathbb{E}}^{ub}_{u,e} - \hat{\mathbb{E}}^{lb}_{v,e}]$, where $\hat{\mathbb{E}}^{ub}_{u,e}, \hat{\mathbb{E}}^{lb}_{v,e}$
are computed using a tail-bound (e.g., Hoeffding inequality \citep{hoeffding1994probability}) from the $\hat{\mathbb{E}}_{u,e}, \hat{\mathbb{E}}_{v,e}$ with a confidence interval $\alpha$.

\begin{proposition}
    \label{prop:emp}
    A model $h_{\theta^T}$ optimized by Algorithm \ref{alg:fairdp} satisfies empirical $\tau_{emp}$-fairness certification with $\tau_{emp} = \max_{u, v \in [K]}[\hat{\mathbb{E}}^{ub}_{u,e} - \hat{\mathbb{E}}^{lb}_{v,e}] + \mathcal{O}(N^{-1/2})$ with a broken probability $(1-\alpha)$.
\end{proposition}

The proof of Proposition \ref{prop:emp} is provided in Appx. \ref{appx:proofs}. In our experiments, we use the Hoeffding inequality with $\alpha=0.95$. 

\textbf{Remark 2.} The Proposition \ref{prop:emp} can be used as a signal for the service provider to monitor the training process. Specifically, the service provider can train the model with \textsc{FairDP} until the empirical fairness certification is larger than a targeted threshold. Then, the service provider can halt the training process. In case the service providers publish the model with the fairness certification, they can leverage the Laplace mechanism \citep{dwork2014} to add Laplace noise $Lap(\frac{1}{\epsilon'})$ to $\hat{\mathbb{E}}_{k,e}$ with extra privacy budget $\epsilon'$ to provide protection for releasing the fairness certification.

\section{Convergence Analysis and Utility Loss Bound}
\label{appx:conv_analysis}

This section focuses on deriving a utility loss bound in terms of convergence analysis for \textsc{FairDP} to provide guidelines on applying the mechanism to downstream tasks. The key observation is that, given a fixed noise scale $\sigma$ and a careful decay of the learning rate $\eta$, \textsc{FairDP} will converge to the global minima for a convex and $\beta$-Lipschitz empirical loss function $\mathcal{L}(D)$ with the rate of $\mathcal{O}(\log T/\sqrt{T})$. To derive such guarantees, we consider the two following assumptions:
\begin{assumption}
    $\mathcal{L}(D, \theta)$ is a convex function with respect to $\theta$, i.e., $\forall \theta, \theta': 
    \mathcal{L}(D, \theta) - \mathcal{L}(D, \theta') \le \langle\nabla_\theta\mathcal{L}(D, \theta), \theta - \theta'\rangle$.
\end{assumption}
\begin{assumption}
    $\mathcal{L}(D, \theta)$ is a $\beta$-Lipschitz function with respect to $\theta$, i.e., $\forall \theta, \theta': 
    |\mathcal{L}(D, \theta) - \mathcal{L}(D, \theta')| \le \beta\|\theta - \theta'\|_2$.  
\end{assumption}
\noindent These two assumptions are generally considered in previous works for private stochastic gradient descent \citep{bassily2014private,bassily19,feldman2020private}. Moreover, they are practically common for many ML models such as Linear Regression, Logistic Regression, and simple neural networks \citep{pilanci2020neural}.

Given these assumptions, we bound the expected empirical risk of a model $h_{\theta^T}$ trained by \textsc{FairDP} as follows:
\begin{theorem}
    \label{theo:utility_loss}
    Let $\theta^T$ be the output of Algorithm \ref{alg:fairdp}. If $C$ is chosen such that $C \ge \beta$, $\mathcal{L}$ satisfies the considered assumptions, and the learning rate $\eta(t) = \mathcal{O}\Big(\frac{K}{\sqrt{t(\beta^2 + K\sigma^2C^2r)}}\Big)$, then the excessive risk $\mathbb{E}[\mathcal{L}(D, \theta^T)] - \mathcal{L}(D, \theta^*)$ is bounded by:
    \begin{align}
        \mathbb{E}[\mathcal{L}(D, \theta^T)] - \mathcal{L}(D, \theta^*) \le \mathcal{O}\Big(\frac{\sigma C\sqrt{r}\log(T)}{\sqrt{TK}}\Big)
    \end{align}
    where $\theta^* = \arg\min_\theta \mathcal{L}(D, \theta)$, $r$ is the number of parameters in $\theta$, and the expectation is over the randomness of \textsc{FairDP}.
\end{theorem}

\begin{proof}
    Considering an updating step $t+1$ of \textsc{FairDP}
    \begin{align}
        &\forall i \in [K]: \theta_i^{t+1} = \theta^t - \eta_t\tilde{g}_i^{t}, \text{ where } \tilde{g}_i^{t} = \bar{g}_i^{t} + \sigma C \mathcal{N}(0, I_r) \nonumber\\
        &\theta^{t+1} = \frac{1}{K}\sum_{i=1}^{K}\theta_i^{t+1}  = \theta^t - \frac{\eta_t}{K}\sum_{i=1}^{K}\bar{g}_i^t + \frac{\eta_t}{\sqrt{K}}\sigma C\mathcal{N}(0, I_r) \nonumber
    \end{align}
    
    \noindent Denote $\tilde{G}_t = \sum_{i=1}^{K}\bar{g}_i^t + \sqrt{K}\sigma C\mathcal{N}(0, I_r)$ and $\bar{G}_t = \sum_{i=1}^{K}\bar{g}_i^t$, then $\mathbb{E}_{\sim \mathcal{N}(0, I_r)}(\tilde{G}_t) = \bar{G}_t$. Furthermore, if $C$ is chosen such that $C \ge \beta$, then $\mathbb{E}(\bar{G}_t) = \nabla_{\theta^t}\mathcal{L}(D, \theta^t)$ where the randomness is over the sampling process. Moreover, we can upper bound the following expectation:
    
    \begin{align}
        \mathbb{E}(\|\tilde{G}_t\|_2^2) &= \mathbb{E}(\|\bar{G}_t + \sqrt{K}\sigma C\mathcal{N}(0, I_r)\|_2^2) \nonumber\\
        &= \mathbb{E}(\|\bar{G}_t\|_2^2) + 2\mathbb{E}(\langle\tilde{G}_t, \sqrt{K}\sigma C\mathcal{N}(0, I_r)\rangle) \nonumber\\
        &+ K\sigma^2 C^2\mathbb{E}(\|\mathcal{N}(0, I_r)\|_2^2) \nonumber\\
        &\le m^2\beta^2 + K\sigma^2 C^2 r = G^2 \label{eq:expectgradnormbound}
    \end{align}
    
    \noindent Then, we can leverage the result from Theorem 2 from \citep{shamir2013stochastic} which declares as follows: 
    \begin{theorem}[Theorem 2 of \citep{shamir2013stochastic}]
        \label{theo:conv_rate}
        For a convex function $\mathcal{L}(\theta)$, let $\theta \in \Theta$ such that $\|\Theta\|_2 \le Q$, $\theta^* = \arg\min_{\theta}\mathcal{L}(\theta)$, and $\theta^0$ is an arbitrary point in $\Theta$. Consider a stochastic gradient descent with $\mathbb{E}(\tilde{G}_t) = \nabla_\theta^t\mathcal{L}$ and the learning rate $\eta_t = \frac{Q}{G\sqrt{t}}$. Then for any $T > 1$, the following is true
        \begin{align}
            \mathbb{E}(\mathcal{L}(D, \theta^T)) - \mathcal{L}(D, \theta^*) \le \mathcal{O}(\frac{QG\log(T)}{\sqrt{T}})
        \end{align}
    \end{theorem}
    \noindent Using the chosen form of $\eta_t$, the bound in Eq. \eqref{eq:expectgradnormbound} with Theorem \ref{theo:conv_rate}, we can derive the guarantee in Theorem \ref{theo:utility_loss} which concludes the proof.
\end{proof}

Theorem \ref{theo:utility_loss} provides guidance on choosing the hyper-parameters $C$ to optimize the convergence rate of \textsc{FairDP}. A larger value of the gradient clipping $C$ will preserve the direction of the clean gradients but also increase the variance, which requires more updating steps to reach convergence. In addition, for simple ML models whose Lipschitz constant $\beta$ can be calculated or approximated, the practitioners can choose $C = \beta$ to maintain the gradient's direction under the clipping process while avoiding excessive DP-preserving noise. 

Practitioners can leverage our results to better balance the trade-offs among privacy, fairness, and utility by adaptively adjusting the training process of \textsc{FairDP}. For example, applying optimizers like Adam \citep{kingma2014adam} at the onset of training may enhance model utility and convergence rate under the same DP protection. As the model nears convergence, practitioners can transition to SGD to secure fairness certification, enabling us to overcome tight constraints on the weights of the last layer. Also, practitioners can adjust the hyper-parameter $M$ to achieve better fairness, such that the smaller $M$, the fairer the model is. However, small $M$ could degrade model utility since it constrains the decision boundary in smaller parameter~space. %(see Table \ref{tab:impact-M} for details)

To our knowledge, \textsc{FairDP} is the first mechanism that achieves $\tau$-fairness certification while preserving DP, without undue sacrificing model utility, as demonstrated in experimental results below. Theorem \ref{theo:worstbound} and Proposition \ref{prop:emp} provide an insightful understanding of the interplay between privacy and fairness.
A stronger privacy guarantee (larger noise scale $\sigma$) tends to result in better fairness certification (smaller $\tau$). In addition, another application of Proposition \ref{prop:emp} is to train a model achieving desirable privacy and fairness guarantees $(\epsilon, \tau)$, which can be predefined by practitioners, by training the model until privacy and empirical fairness estimate aligns with the predetermined thresholds and halting the training if one of the guarantees is breached.

\section{Experimental Results}

We conducted a comprehensive evaluation of \textsc{FairDP} and baseline methods on various benchmark datasets, primarily focusing on two aspects: \textbf{(1)} Assessing the accuracy and tightness of the fairness certification by comparing it with empirical results obtained from multiple statistical fairness metrics; \textbf{(2)} Examining the trade-off between model utility, privacy, and fairness; and \textbf{(3)} Exploring the contribution of each component of \textsc{FairDP} to the overall utility and fairness of the models through extensive ablation studies.

% \subsection{Datasets, Metrics, and Model Configurations}

\textbf{Datasets, Metrics, and Model Configurations.} The evaluation uses three datasets: the Adult dataset \citep{uci}, the Default of Credit Card Clients (Default-CCC) dataset \citep{yeh2009comparisons}, and the UTK-Face Dataset (UTK) \citep{zhifei2017cvpr}. Details of the datasets are in Table \ref{tab:dataset}. 
These are the benchmark datasets to evaluate the fairness-aware ML algorithm \citep{tran2022pruning, icml2022, han2023ffb}.
Data preprocessing steps are strictly followed as outlined in previous works such as \citep{iofinova2021flea, ruoss2020learning,tran2021differentially, han2023ffb}. Since the datasets are extremely imbalanced (i.e., the number of positive data is much smaller than the number of negative data), we evaluate the model's utility by using \textit{area under the ROC curve} (ROC-AUC) and Accuracy as in previous studies.
% \Linh{Why are they extremely imbalanced? Any impact on the results between imbalanced vs. balanced datasets? Also, I'd reverse the sentence, i.e., putting `Since the datasets are...' first.}
A {\it higher} ROC-AUC (Accuracy) indicates {\it better} utility. We use \emph{demographic parity} \citep{dwork2012fairness}, \emph{equality of opportunity}, and \emph{equality of odds} \citep{hardt2016equality} as primary fairness metrics since they are the standard metrics for fairness measurement.
% \Linh{Justify the metrics!} 
The lower values of these fairness metrics indicate fairer decisions. The experiments use privacy budgets in the range of $[0.5, 2.0]$ and $\delta=1e^{-5}$ for different datasets. Although DP is celebrated for using small values of $\epsilon$, most  current deployments report $\epsilon$ larger than $1$, with many of them using $\epsilon > 5$ \citep{Desfontain}.
% \footnote{\url{https://desfontain.es/privacy/real-world-differential-privacy.html}}  
% Therefore, since fairness is affected by privacy loss, this work is important to highlight and justify the trade-offs between privacy and fairness within this privacy loss regime.
% \Linh{By `this work', do you mean the paper, the choice of $\epsilon$, or the evaluation? The phrase is a little unclear.}

In Adult and Default-CCC datasets, a multi-layer perceptron (MLP) is employed with ReLU activation on hidden layers and sigmoid activation on the last layer for binary classification tasks. For the UTK dataset, a simple Convolution Neural Network (CNN) is employed since it is an image-based dataset as considered in \citep{tran2022pruning}.
% \Linh{Not sure why the terms "On the one hand" and "On the other hand" are needed. You just describe the characteristics of each dataset, not consider both types together here.} 
Adam optimizer \citep{kingma2014adam} is employed during the complete training process.
For \textsc{FairDP}, we set the weight clipping hyper-parameter $M \in [0.1, 1.0]$ and initialize the learning rate $\eta = 0.007$ for the first half of the training process, and then reduce it to $\eta=0.005$ for the rest of the process. Statistical tests used are two-tailed t-tests.

\begin{table}[t]
    \caption{Evaluation Datasets.}
    \label{tab:dataset}
    \begin{center}
    \resizebox{\columnwidth}{!}{%
    \begin{tabular}{lccc}
    \toprule
    Dataset & Default-CCC & Adult & UTK \\
    \midrule
    \# data  & 30,000 & 48,842 & 23,795 \\
    \# features & 89 & 41 & 48 $\times$ 48 $\times$ 1\\
    \# positive & 6,636 & 11,687 & 8,608\\
    protected attribute & Gender & Gender & Race \\
    \bottomrule
    \end{tabular}
    }
    \end{center}
    \vspace{-10pt}
\end{table}

\textbf{Baselines.}
We consider a variety of DP-preserving mechanisms, fairness training algorithms, and combinations of these as baselines, resulting in six baselines, including a non-DP mechanism, three existing mechanisms that either preserve DP or promote fairness, one existing mechanism that achieves both fairness and privacy, and one adapted mechanism that achieves both DP and fairness. 
% and two variants of \textsc{FairDP}.

\textbf{Established Baselines.} We consider \textbf{DP-SGD} \citep{abadi:16}, \textbf{DPSGDF} \citep{xu2021removing}, \textbf{FairSmooth} \citep{icml2022}, and \textbf{DP-IS-SGD} \citep{kulynych2022you} as baselines. \textbf{DP-SGD} is a well-established DP mechanism with many applications in DP research.
\textbf{DPSGDF} is designed to alleviate the disparate impact of DPSGD by focusing on accuracy parity. \textbf{FairSmooth} is a state-of-the-art mechanism that assures group fairness by transforming the model $h_\theta$ into a smooth classifier as $\hat{h}_{\theta} = \mathbb{E}_{\nu}[h_{(\theta + \nu)}]$ where $\nu \sim \mathcal{N}(0, \bar{\sigma}^2)$ in the inference process, where $\bar{\sigma}$ is the standard deviation of the Gaussian noise. \textbf{DP-IS-SGD} established the connection between DP and distributional generalization and reduced the accuracy disparity through important sampling during the training process. In addition, we also consider a combination of the baselines, \textbf{DPSGD-Smooth}, by applying \textbf{FairSmooth} to models trained by \textbf{DPSGD}.  
Since it is the only baseline offering both DP and fairness guarantees, we employ it for comparison against \textsc{FairDP}. 
We also acknowledge a previous work DP-FERMI \citep{lowy2023stochastic}; however, we do not consider it as a baseline in this study since the implementation of DP-FERMI has not been finalized per indicated by the~authors.

\begin{figure*}[h]
    \centering
    \includegraphics[width=
    1.92\columnwidth]{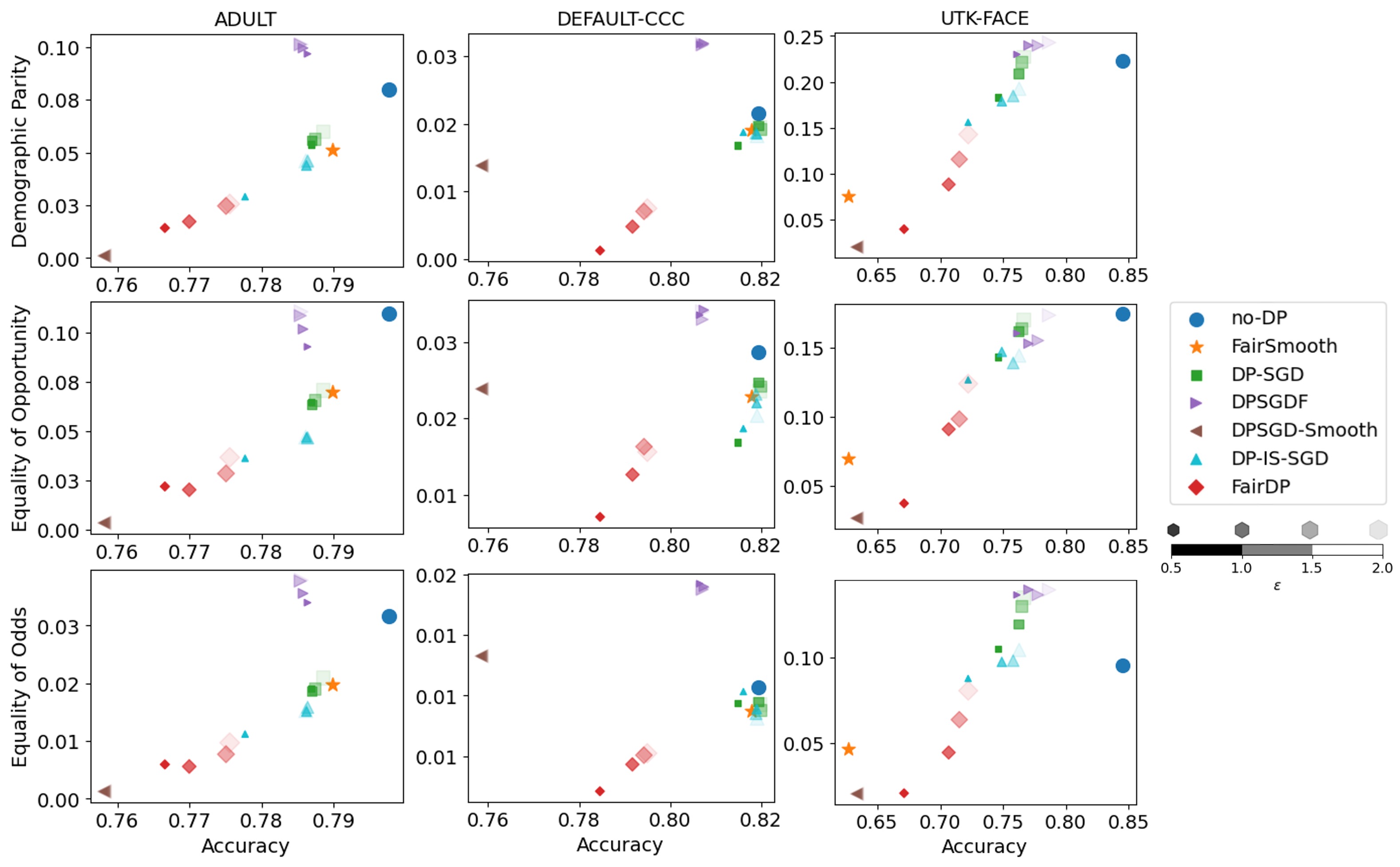}\vspace{-1ex}
    \caption{Trade-off among model utility (Accuracy), DP-preservation, and fairness.}
    \label{fig:alljoint-full-acc}
    \vspace{-1ex}
\end{figure*}

% \hai{Include the baseline that is incorrect, fix it, and run its experiments.}

% $\bullet$ \textbf{Variants of \textsc{FairDP}.} 
% To examine how different features of \textsc{FairDP} affect the model performance and fairness, we introduce two \textsc{FairDP} variants, called \textbf{FairFM} and \textbf{FairFM-Smooth}. 
% \textbf{FairFM} (refer to Appx. \ref{appex:alg}) distinguishes itself from \textsc{FairDP} by incorporating noise into the objective function as a pre-processing step to preserve DP. The \textbf{FairFM-Smooth} mechanism is a variant of FairFM that applies the FairSmooth method \citep{icml2022} to the model trained by FairFM during the inference process.  

\subsection{Utility, Privacy, and Fairness Trade-offs} 

\begin{figure*}[h]
    \centering
    \includegraphics[width=
    1.92\columnwidth]{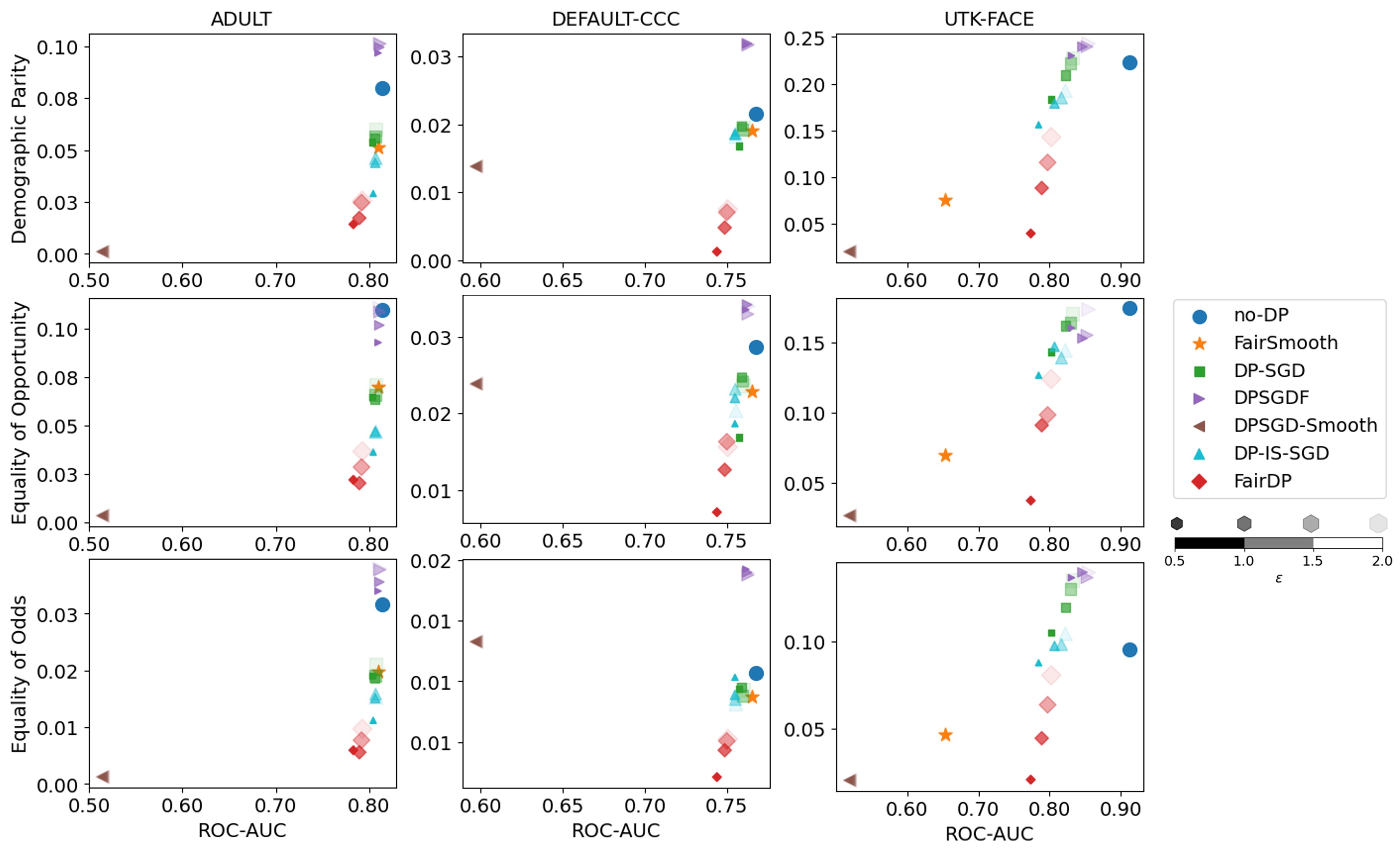}\vspace{-1ex}
    \caption{Trade-off among model utility (ROC-AUC), DP-preservation, and fairness.}
    \label{fig:alljoint-full-auc}
    \vspace{-1ex}
\end{figure*}

Fig. \ref{fig:alljoint-full-acc} and Fig. \ref{fig:alljoint-full-auc} show the results of each algorithm w.r.t. model's utility, fairness, and privacy. In these figures, the points indicate the average results of an experiment of \textsc{FairDP} and baselines with the corresponding privacy budget. The smaller and darker points indicate experiments with lower privacy budgets and vice-versa. In Fig. \ref{fig:alljoint-full-acc} and Fig. \ref{fig:alljoint-full-auc}, points positioned closer to the bottom-right corner denote superior balance among model utilty (higher accuracy/precision), privacy (strict DP protection), and fairness (lower empirical values of fairness metrics). 

\textbf{Comparing with Baselines}. In general, \textsc{FairDP} significantly attains fairer decisions while maintaining high utility across datasets and fairness metrics with different privacy budgets compared with the clean model and the best baseline~DP-IS-SGD. 

\begin{figure*}[h]
    \centering
    \includegraphics[width=0.5\columnwidth]{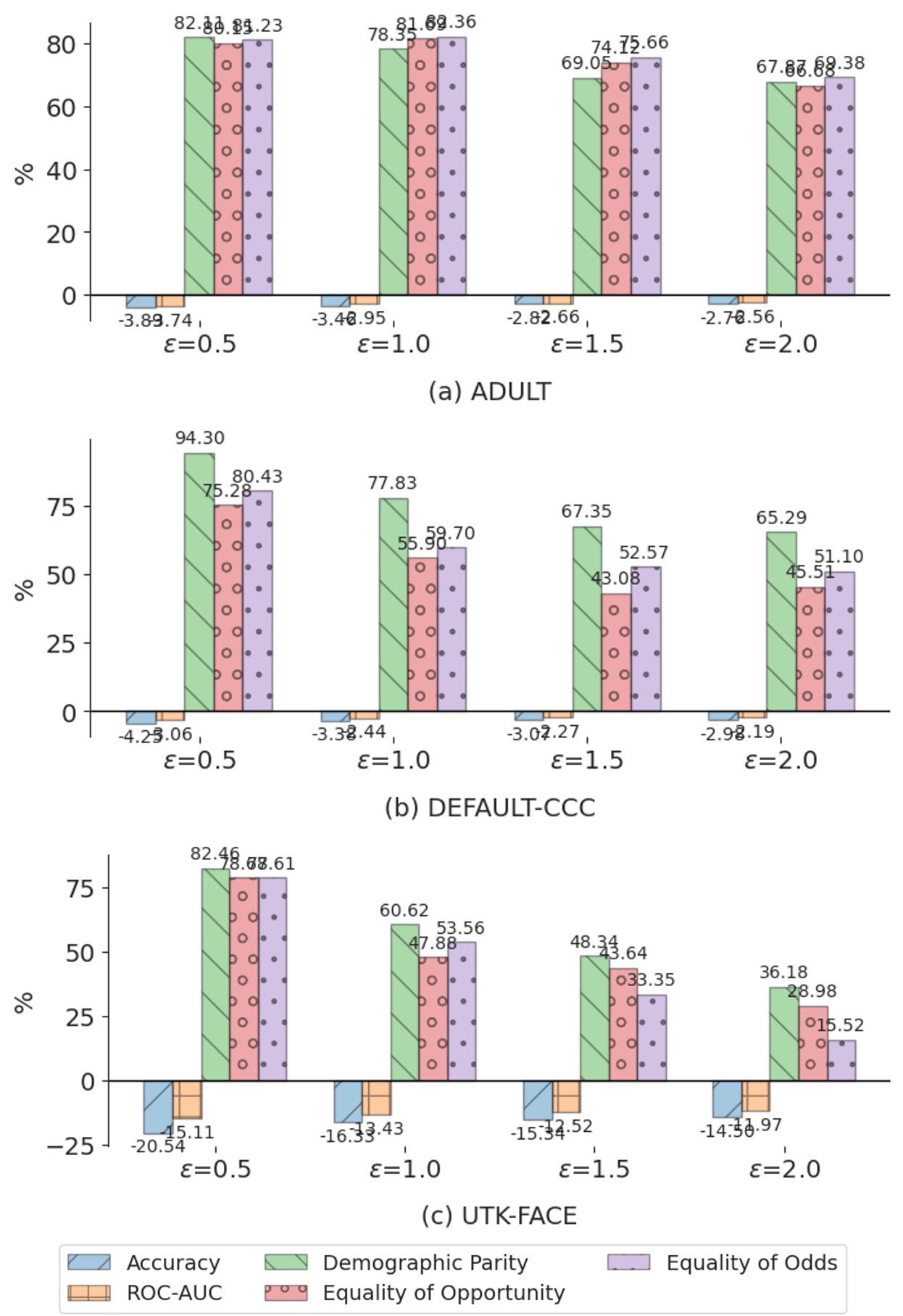}%\vspace{-20pt}
    \caption{Relative utility drop and fairness gain of \textsc{FairDP} compared with the clean model.}
    % \vspace{-8pt}
    \label{fig:reduced-compared-clean}
\end{figure*}

Specifically, in the Adult dataset, under rigorous privacy protection ($\epsilon = 0.5$), \textsc{FairDP} gains 75\% improvement in demographic parity while attaining a modest decrease in ROC-AUC and Accuracy compared to the clean model (3\% in ROC-AUC, 4.3\% in Accuracy on average) average across different privacy budgets ($\epsilon \in [0.5, 2]$). Specifically, at $\epsilon =0.5$, the demographic parity reduce from $0.079$ in the clean model to $0.014$ of \textsc{FairDP} with $p$-value $=3.15e^{-5}$. Fig. \ref{fig:reduced-compared-clean} (Appx. \ref{Appx:supp}) illustrates the comparison between \textsc{FairDP} and the clean model in terms of fairness gains compared to utility drops, which shows that \textsc{FairDP} achieves more than $60\%$ improvement in terms of fairness while attaining marginal drop of utility (i.e., less than $5$\%) across different privacy budgets. Similar results between \textsc{FairDP} and the clean model are observed for other datasets and other fairness metrics with different privacy~budgets.

% Similarly, \textsc{FairDP} achieves significant improvement in terms of fairness (??\% across fairness metrics on average) compared with the best baseline DP-IS-SGD while registering a marginal drop in terms of ROC-AUC and Accuracy (i.e., less than $3\%$ and $6\%$ on average) across different privacy budgets ($\epsilon \in [0.5, 2]$). Specifically, 

\begin{figure}[t]
    \centering
    \includegraphics[width=0.9\columnwidth]{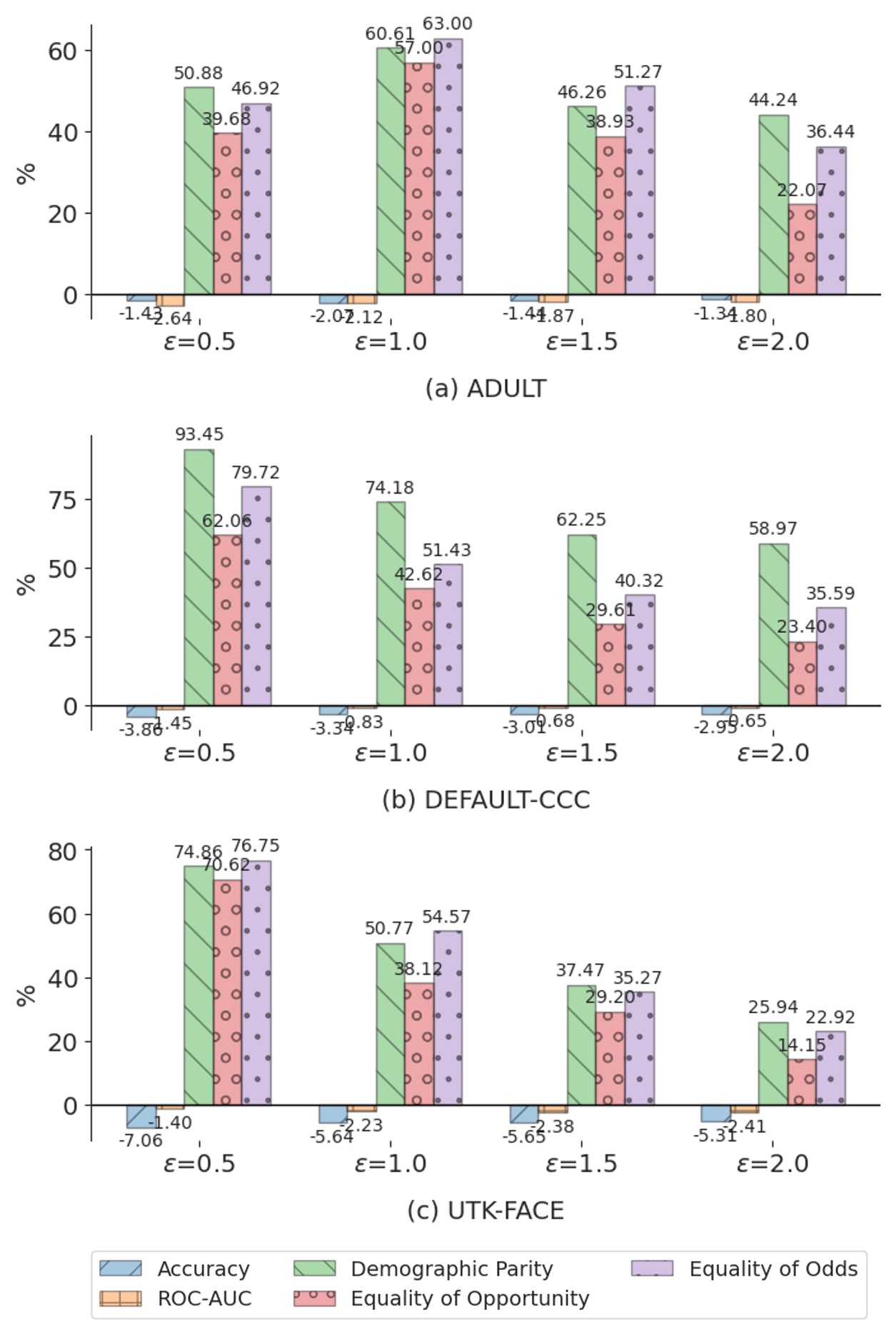}%\vspace{-20pt}
    \caption{Relative utility drop and fairness gain of \textsc{FairDP} compared with the best baseline - DP-IS-SGD.}
    % \vspace{-8pt}
    \label{fig:reduced-compared}
\end{figure}

Fig. \ref{fig:reduced-compared} illustrates the comparison between \textsc{FairDP} and the best baseline (DP-IS-SGD) in terms of fairness gains and utility drops. At $\epsilon=0.5$, \textsc{FairDP} achieves a significant gain across fairness metrics (over $45.83\%$) compared with DP-IS-SGD while attaining marginal utility drop (less than $2.64\%$) for the Adult dataset. Furthermore, \textsc{FairDP} attains $50.5\%$ gain on average in terms of demographic parity for the Adult dataset across different privacy budgets. In addition, under looser DP protection (i.e., $\epsilon=2.0$) where \textsc{FairDP} is less fair, \textsc{FairDP} still achieves a significant fairness gain ($34\%$) compared with DP-IS-SGD while maintaining marginal drop of utility (less than $1.8\%$) on average across fairness and utility metrics. Similar results are observed in other datasets and other fairness metrics. In fact, the fairness gains in \textsc{FairDP} go up to 78.41\% and 74.08\% on average across fairness metrics in the Default-CCC and UTK-Face datasets given $\epsilon = 0.5$ correspondingly.

These results highlight the effectiveness of \textsc{FairDP} in addressing the trade-off among privacy, fairness, and utility for different ML tasks. 
\textit{The promising results of \textsc{FairDP} can be attributed to its unique approach of controlling the contribution of each group to the learning process, the DP-preserving noise injected into each group, enforcing a constraint on the decision boundary, and aggregating the knowledge learned from every group at each training step. \textsc{FairDP} fundamentally differs from the baselines, leading to its superior performance.}

% \end{minipage}}
Another noteworthy observation is that treating fairness as a constraint, as in the case of DPSGDF, does not consistently improve the trade-offs among model utility, privacy, and fairness. In the Adult dataset, DPSGDF is less fair than the clean model in terms of demographic parity ($0.1$ compared with $0.079$ in demographic parity with $p = 3.84e^{-7}$, i.e., $25\%$ increase). This can be attributed to the fact that handling all groups simultaneously within the noisy SGD process can hide the information from minor groups, leading to a degradation in fairness. Also, the fairness constraints, employed as penalty functions, have an impact on the optimization of the model, leading to a deterioration in its performance for ML tasks.

% \begin{figure}[t]
%     \centering
%     \includegraphics[width=0.85\columnwidth]{images/reduced-demo-all.jpg} %\vspace{-25pt}
%     \caption{Trade-off among model utility, privacy-budget, and demographic parity. The results for Accuracy and other fairness metrics are provided in Fig. \ref{fig:alljoint-full-auc} and Fig. \ref{fig:alljoint-full-acc}(Appx. \ref{Appx:supp})}
%     \label{fig:reduced-dp}
% \end{figure}

These issues can be mitigated by separating the DP-preserving training process from the methods developed to attain fairness during inference, as in the case of DPSGD-Smooth. These methods achieve better $\tau$-fairness with relatively competitive model utility under equivalent DP protection. However, this approach does not effectively balance the trade-offs among model utility, privacy, and fairness as effectively as \textsc{FairDP} does. 
{\it These insights highlight the need to explore novel approaches to seamlessly integrate DP-preserving and fairness rather than treating them as independent (constrained) components. \textsc{FairDP} represents a pioneering step in this direction.}

\textbf{Privacy and Fairness Correlation.} Our observation from the results is that \textit{privacy enhances fairness}. Specifically, the lower the value of the privacy budget $\epsilon$ (stronger privacy protection), the lower the fairness metric (Fig. \ref{fig:alljoint-full-acc}) and the higher the fairness gain (Fig. \ref{fig:reduced-compared}). For instance, in the Adult dataset, the demographic parity gains $44\%$ (i.e., reducing from $0.025$ to $0.014$) when the privacy budget decreases from $\epsilon=2.0$ to $\epsilon=0.5$. Similar behavior is observed for other datasets and other fairness metrics. This observation verifies the validity of Remark 1 and our fairness certification. {\it These insights highlight the contribution of \textsc{FairDP} in understanding the correlation between fairness and privacy.}

% \textcolor{blue}{Issa: An important point that is missing here is that Figure 4 shows that the fairness gain increases with increasing privacy levels which contradicts the remark you had in the introduction (I commented on it there). The figure shows the more intuitive conclusion that privacy enhances fairness, not the other way around, that is, privacy and fairness are not conflicting requirements.} 

% \hai{Addressing the comment from Dr.Issa here.}

\subsection{Fairness Certification under Data Distribution Shift}

\begin{figure}[t]
    \centering
    \includegraphics[width=
    0.9\columnwidth]{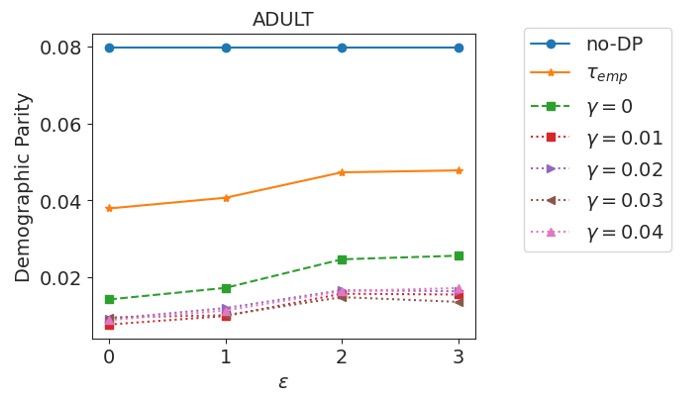}
    %\vspace{-15pt}
    \caption{Tightness of empirical fairness certification on demographic parity w.r.t different privacy budgets for the Adult dataset. 
    The results for other fairness metrics and datasets are in Fig. \ref{fig:bddist-full}, Appx. \ref{Appx:supp}} % \vspace{-3ex}
    \label{fig:tau_emp}
\end{figure}

\begin{figure*}[h]
    \centering
    \includegraphics[width=
    1\columnwidth]{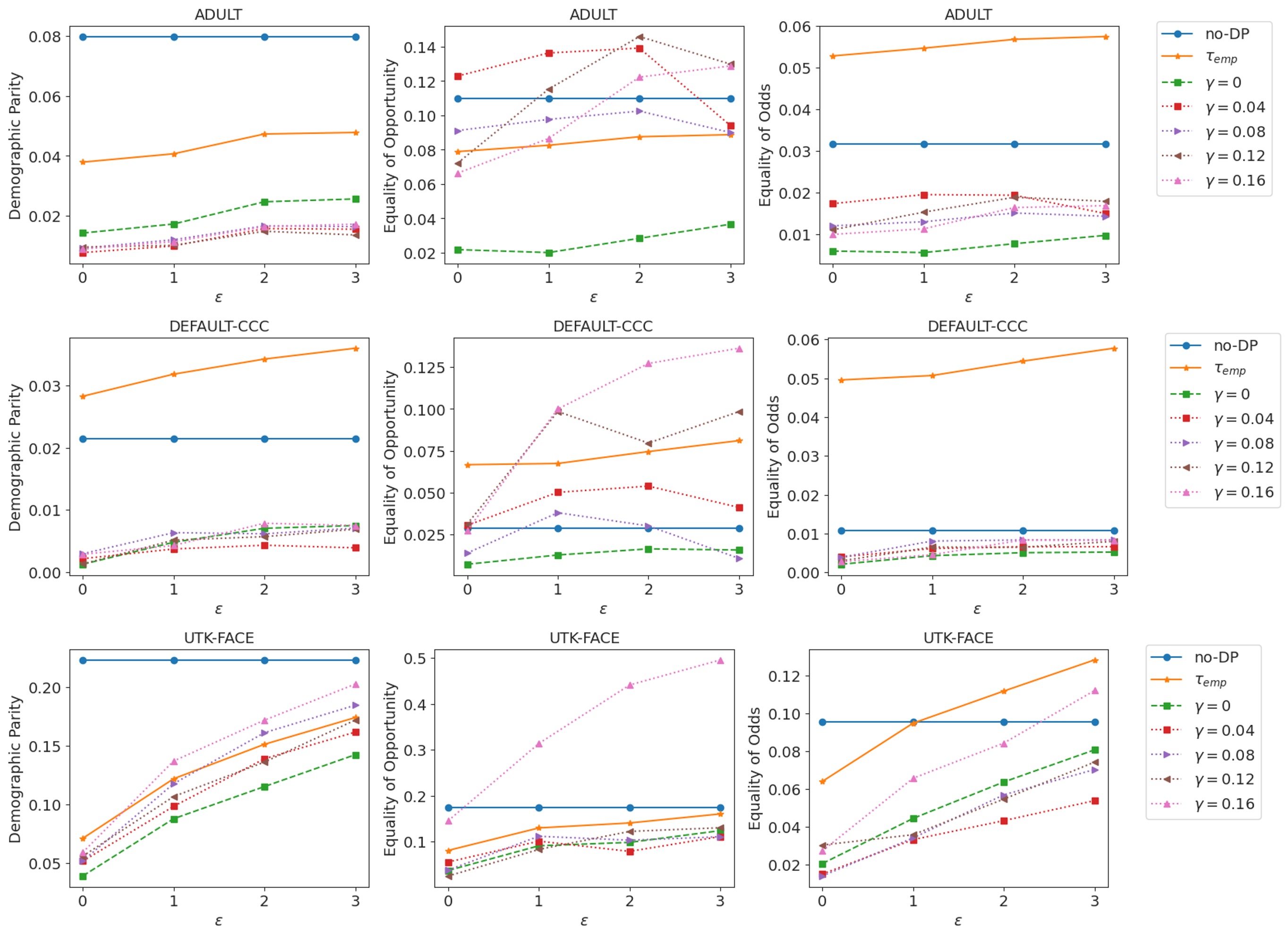}
    \caption{Empirical fairness certification under distribution shift. $\gamma$ is the total variance between $Pr(y, a)$ of the train and test sets.}
    \label{fig:bddist-full}
\end{figure*}

\textbf{Tightness of Fairness Certification.} Fig. \ref{fig:tau_emp} and \ref{fig:bddist-full} (Appx. \ref{Appx:supp}) show the empirical fairness results and the certification $\tau_{emp}$. It is worth noting that we inject the Laplace noise with $\epsilon' = 0.1$ into the empirical fairness certification in our experiments. Although we make an assumption that the training data and the testing data are from the same distribution, we create a shift in the two distributions quantified by $\gamma$, which measures the total variation between the joint probability $Pr(a, y)$ in the training and testing datasets. To create the distribution shift, we modify the joint probability $Pr(a, y)$ as described in \citep{an2022transferring}. In general, the empirical results confirm the validity of our $\tau_{emp}$-fairness certification across different datasets and privacy budgets since the empirical value of the fairness metrics ($\gamma=0$) is lower than the certification. Furthermore, in most instances for the Adult and UTK-Face datasets, our $\tau_{emp}$-fairness certifications are substantially lower than the empirical fairness values of the clean model.
For instance, for demographic parity in the UTK-Face dataset, our empirical certifications are significantly smaller than the empirical fairness results of the clean model ($p=2.14e^{-5}$) while maintaining a small gap with the empirical fairness results of \textsc{FairDP}. Moreover, across the different values of $\gamma$, we can observe that our $\tau_{emp}$-fairness certification still holds under some magnitude distribution shift, which highlights the robustness of the certification in real-world scenarios. That shows the correctness and tightness of our $\tau_{emp}$-fairness certification across datasets and DP budgets, strengthening the advantages of \textsc{FairDP} in theoretical guarantees and empirical results compared with the baselines. Note that $\tau_{emp}$ can be marginally larger than the empirical fairness results of the clean model for some instances. It is because our certification is tailored to DP preservation in \textsc{FairDP} instead of the (non-DP-preserving) clean model.

% , but it's a certification based on the mechanism of \textsc{FairDP}. However, practitioners can still leverage our certification to fine-tune their models to achieve the desired fairness level. We will improve our certification by tailoring it toward the clean model in the future works.}

\textbf{DP-MC Approximation Error.} Table \ref{tab:monte_error} shows the error of the DP-MC approximation when $N=10$. In general, the Monte Carlo error is small across the datasets compared to the $\tau_{emp}$-fairness certification. Specifically, in the Adult dataset, the error is $4.37e^{-5}$, which is $0.1\%$ of the $\tau_{emp}$-fairness certification for demographic parity across different privacy budgets. Similar results are observed for other datasets and fairness metrics, which highlights the practicality of \textsc{FairDP} when collaborating the $\tau$-fairness certification with the DP-MC approximation, strengthening the advantages of \textsc{FairDP} in theoretical guarantees.

\begin{table}[h]
    \caption{Monte Carlo approximation error of fairness certification with $N=10$ across different datasets.}
    \label{tab:monte_error}
    \begin{center}
        % \begin{sc}
        \begin{small}
        \resizebox{\columnwidth}{!}{%
        
        \begin{tabular}{lp{5em}p{5em}p{5em}}
        \toprule
        Dataset & Demographic Parity & Equality of Opportunity &  Equality of Odds \\
        \midrule
        Adult & $4.37e^{-5}$ & $3.16e^{-4}$ & $3.7e^{-4}$ \\
        \midrule
        Default-CCC & $1.1e^{-4}$ & $4.8e^{-4}$ &  $6.2e^{-4}$ \\
        \midrule
        UTK & $1.36e^{-4}$ & $3.7e^{-4}$ & $6.1e^{-4}$ \\
        \bottomrule
        \end{tabular}
        
        % \end{sc}
        }
        \end{small}
    \end{center}
\end{table}

\subsection{Ablation Study} 

% \hai{We should mention about Accuracy.}\khang{I already mentioned it in the discussion}

\begin{figure}[t]
    \centering % \vspace{-5pt}
    \includegraphics[width=\columnwidth]{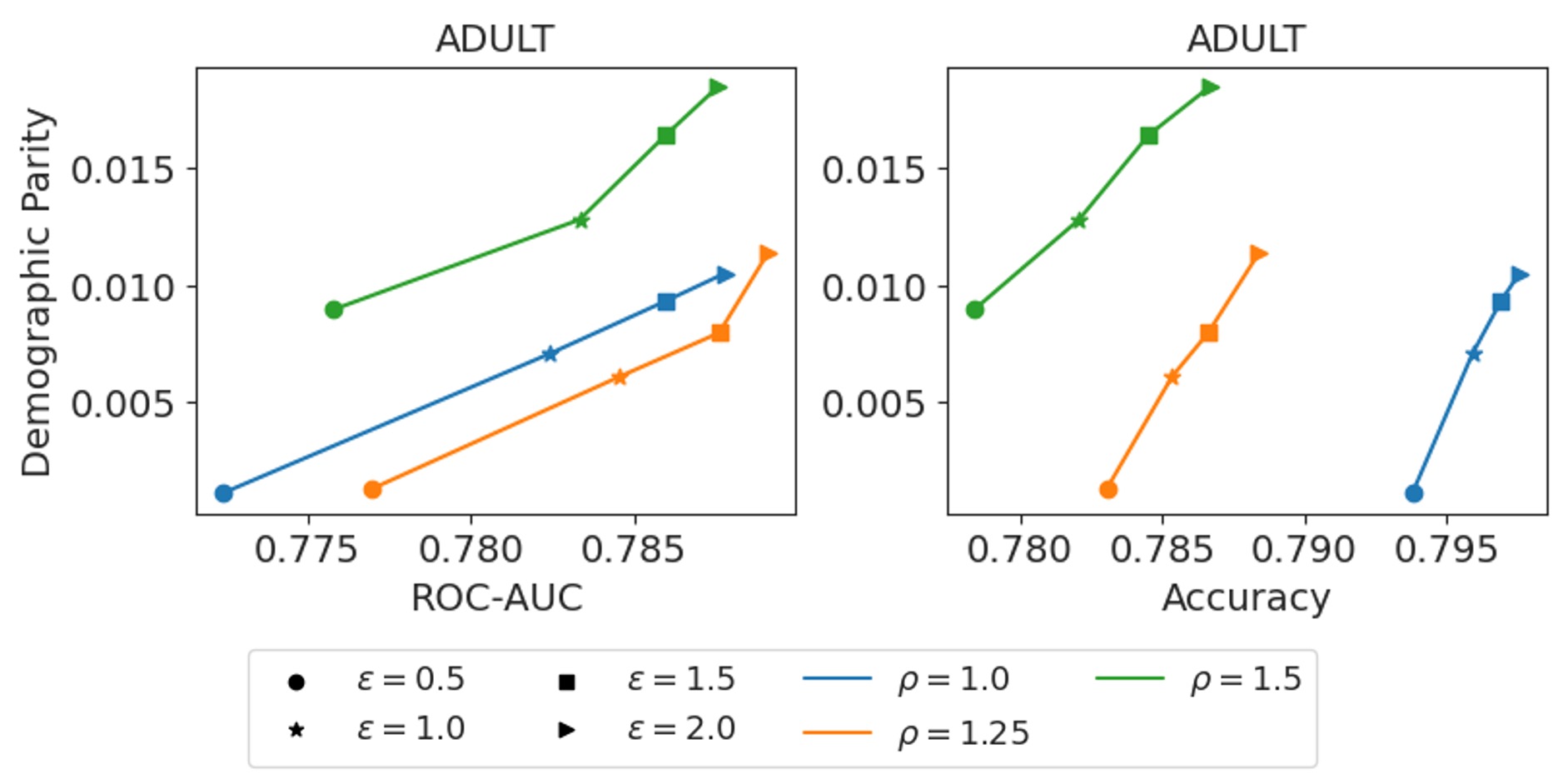} %\vspace{-15pt}
    \caption[l]{Model utility, privacy, and demographic parity for various $\rho$ values on the Adult dataset. The results for other fairness metrics and datasets are in Figs. \ref{fig:rho-full-auc}-\ref{fig:rho-full-acc}, Appx. \ref{Appx:supp}} %\vspace{-10pt}
    \label{fig:adult-rho}
\end{figure}

\begin{figure*}[h]
    \centering
    \includegraphics[width=
    1\columnwidth]{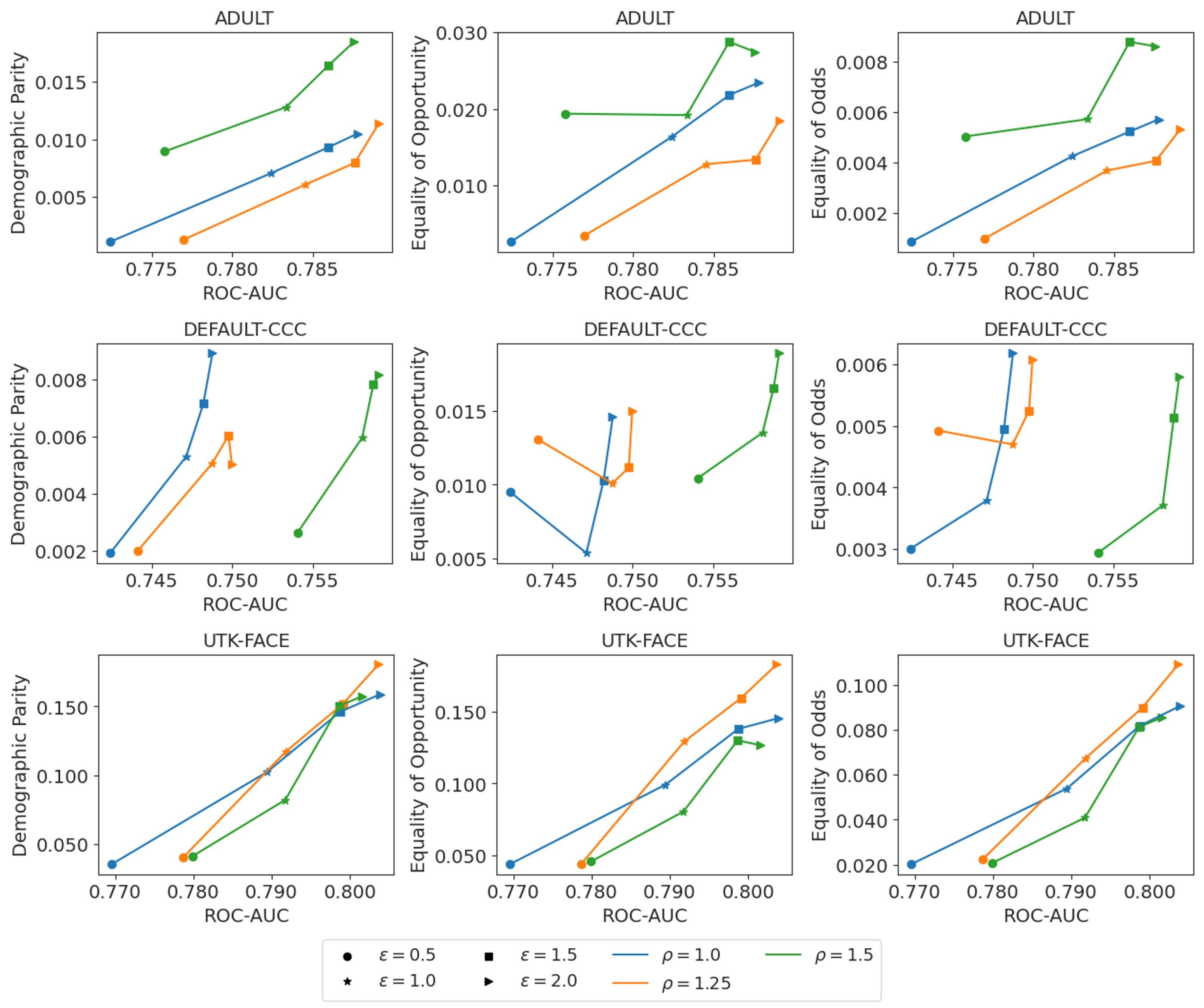}
    \caption{Model utility (ROC-AUC), privacy, and fairness for various $\rho$ values, which measure the ratio between the number of data points of advantage and disadvantage groups.}
    \label{fig:rho-full-auc}
\end{figure*}

\begin{figure*}[h]
    \centering
    \includegraphics[width=
    1\columnwidth]{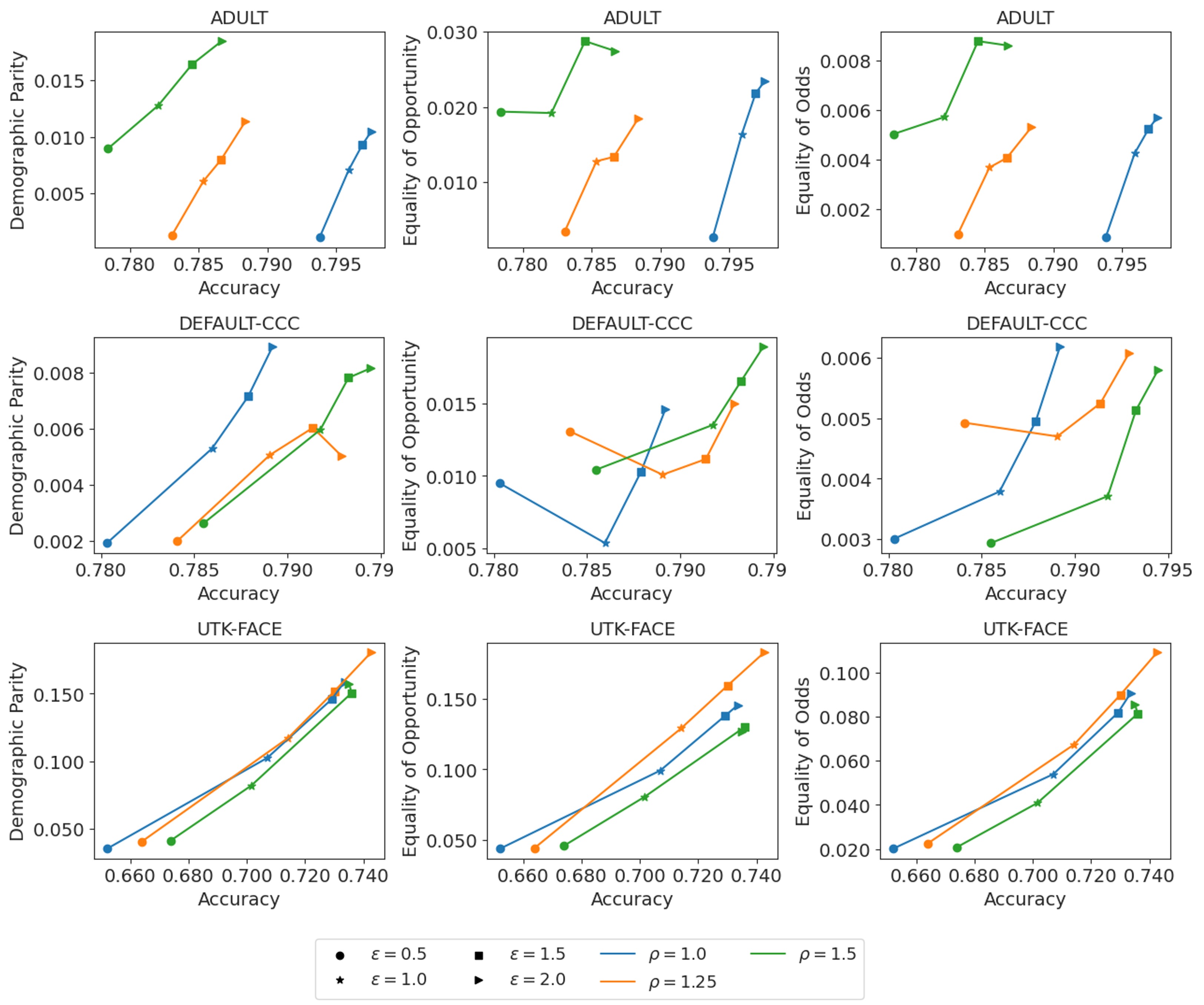}
    \caption{Model utility (Accuracy), privacy, and fairness for various $\rho$ values, which measure the ratio between the number of data points of advantage and disadvantage groups.}
    \label{fig:rho-full-acc}
\end{figure*}

\textbf{Imbalanced Protected Group.} Practitioners can tune \textsc{FairDP} to find an appropriate setting that balances the level of DP protection with the desired level of fairness and model utility. Fig. \ref{fig:adult-rho} and Figs. \ref{fig:rho-full-auc}-\ref{fig:rho-full-acc} (Appx. \ref{Appx:supp}) illustrate the effect of the ratio $\rho$ between the size of the datasets of the minor and major groups: $\rho =  (\arg\max_{a\in[K]}n_a)/(\arg\min_{b\in[K]}n_b)$. For a specific $\rho$, we randomly sample data points from the majority group, reducing the size of the major group to the desired $\rho$. In general, increasing $\rho$ values leads to more data points from the majority group being utilized for training the model, thereby improving its accuracy. However, the effect on the model's fairness across different fairness metrics is not consistently observed. Nonetheless, our guarantee remains applicable across various degrees of dataset imbalance. Lower privacy budgets (i.e., stronger privacy guarantees) contribute to improved fairness in the model's decisions, strengthening the theoretical certification of \textsc{FairDP}.

\textbf{Hyper-parameter $M$.} Table \ref{tab:impact-M} (Appx. \ref{Appx:supp}) shows the effect of the hyper-parameter $M$ toward the trade-off among privacy, fairness and utility. In general, reducing $M$ leads to a fairer model's decision, as described in our fairness certification. However, reducing $M$ will restrict the model's decision boundary, resulting in the degradation of model utility. In the Adult dataset, when $M$ drops from $1.0$ to $0.25$, the ROC-AUC drops from $0.80$ to $0.59$ ($\approx 50\%$ reduction), and Accuracy drops from $0.80$ to $0.75$ ($\approx 4\%$ reduction), while the demographic parity is reduced from $0.027$ to $0.0$. Practitioners can tune this parameter to better balance the trade-off among privacy, fairness, and utility in \textsc{FairDP}. 

\section{Discussion}
\textsc{FairDP} is a robust framework for integrating differential privacy with group fairness. A fundamental limitation is its assumption that all groups have sufficient data for practical model training, which is often not the case in real-world scenarios where underrepresented groups pose challenges to fairness and utility. However, if data instances of some groups are too small, the practitioner can apply data augmentation techniques \citep{bao2024dp, song2024llm, he2024generative} under privacy protection on top of \textsc{FairDP} to mitigate this limitation. 

In addition, \textsc{FairDP} is that the framework is designed for a centralized setting, where the protected attributes are exposed to the server. In contexts such as hospitals, banks, or universities, the organization typically has access to its members' complete set of protected attributes, which it uses to make informed decisions about privacy and fairness. Nevertheless, applying \textsc{FairDP} to this organizational-level decision-making ensures that privacy and group fairness concerns are addressed in the context of the broader goals and policies of the organization.

Finally, regarding tuning the privacy budgets, practitioners can choose appropriate ones by carefully tailoring them to the specific requirements and sensitivity of their data domain \citep{dwork2019differential}. Then, the practitioner can leverage Theorem 2 and Proposition 3 to tune the noise scale $\sigma$ and clipping values $C, M$ to achieve the desirable fairness and utility.

% Additionally, FairDP’s centralized design leverages organizational control over sensitive attributes to ensure consistent application of fairness criteria. The framework’s flexibility in tuning privacy budgets allows practitioners to balance privacy, fairness, and utility, supported by theoretical foundations like Theorem 2 and Proposition 3. By combining these strategies, practitioners can overcome limitations related to group size and privacy budget selection, enhancing FairDP’s application in diverse settings.

% \vspace{-5pt}

\section{Conclusion}
\label{Conclusion and Discussion}
% \vspace{-5pt}

This paper introduces \textsc{FairDP}, a novel mechanism that achieves certified group 
fairness while preserving DP and sustaining high model utility. Departing from existing mechanisms, the key ideas of \textsc{FairDP} are fairness-aware DP training and Monte Carlo approximation for fairness certification in the inference time, resulting in a rigorous privacy guarantee and fairness certification. Therefore, \textsc{FairDP} provides a comprehensive understanding of the influence of DP-preserving noise on model fairness guarantees and derives tight fairness certification by leveraging the DP-preserving noise. Our extensive experimental results
showed that \textsc{FairDP} enhances the trade-off among model utility, privacy, and fairness, outperforming an array of baselines on benchmark datasets.

\section*{Acknowledgement}
This work is partially supported by grants NSF CNS-1935928, NSF SaTC 2133169, NSF SaTC 1935923, NSF FAI 1939725, NSF SCH 2123809, and QRDI ARG01-0531-230438.

\bibliographystyle{plainnat}
\bibliography{ref}

\newpage
\onecolumn
\appendices

\section{Algorithm}
\label{appx:alg-dpsgd}

\begin{algorithm}[!h]
    \footnotesize
    \caption{DPSGD} \label{alg:dpsgd}
    \begin{algorithmic}[1]
        \STATE \textbf{Input}: Dataset $D$, sampling rate $q$, noise scale $\sigma$, norm bounds $C$ and $M$, number of steps $T$.
        % \State \textbf{Output}: Model's parameter $\theta^{(T)}$
        \STATE Initialize $\theta^{0}$ randomly
        \FOR {$t \in [1:T]$}
        \STATE Sample $B^t$ from $D$ with sampling probability $q$.
        \STATE \textbf{Compute gradient}: For $x_i \in B^t$, $g_i = \nabla_{\theta}\ell(x_i)$
        \STATE \textbf{Clip gradient}: $\bar{g}_i = g_i\min(1, \frac{C}{\|g_i\|_2})$
        \STATE \textbf{Compute total gradient}: $\Delta = \sum_{i \in B^t}\bar{g}_i$
        \STATE \textbf{Add noise}: $\tilde{\Delta}= \Delta + \mathcal{N}(0, C^2\sigma^2I_r)$
        \STATE \textbf{Update}: $\theta^t = \theta^{t-1}- \frac{\eta_t}{|B^t|}\tilde{\Delta}$
        \ENDFOR
        \STATE Return $\theta^{T}$
    \end{algorithmic}
\end{algorithm}

\section{Proofs of Fairness Certification}
\label{appx:proofs}

% \subsection{Proof of Theorem \ref{theo:worstbound}}
% \label{Appx:proof_fair}

\subsection{Proof of Proposition \ref{prop:emp}}

\begin{proof}
    From the considered fairness metric in Eq. \eqref{eq:fairgeneral}, by leveraging Eq. \eqref{eq:distlarger0}, we have that:
    \begin{align}
        \tau &= \max_{u,v} Pr(\hat{y} = 1 | u, e) - Pr(\hat{y} = 1 | v, e) = \max_{u,v}\mathbb{E}_{u,e} - \mathbb{E}_{v,e}
    \end{align}
    where $\mathbb{E}_{k,e} = \mathbb{E}_{x \sim P(x| k, e)}\Big[\frac{1}{2} + \frac{1}{2} \texttt{erf}\Big(\frac{\langle w^{t-1} - \eta\mu^t, \xi\rangle}{\|\xi\|_2\sigma_0\sqrt{2}}\Big)\Big]$. Denote $\hat{\mathbb{E}}_{k,e}$ as the sample mean computed on the training set of group $k$. By leveraging the Central Limit Theorem, there exists an upper-bound $\hat{\mathbb{E}}^{ub}_{u,e} \ge \mathbb{E}_{u,e}$ and a lower-bound $ \hat{\mathbb{E}}^{lb}_{v,e} \le \mathbb{E}_{v,e}$ with the confident interval of $(1-\alpha)$ which can be computed using a tail-bound (e.g., Hoeffding inequality \citep{hoeffding1994probability}). Therefore, along with the MC approximation error, we have that
    \begin{align}
        \tau &= \max_{u,v}\mathbb{E}_{u,e} - \mathbb{E}_{v,e} \le \max_{u,v}\hat{\mathbb{E}}^{ub}_{u,e} - \hat{\mathbb{E}}^{lb}_{v,e} + \mathcal{O}(N^{-1/2})
    \end{align}
    with a confident interval of $(1-\alpha)$, which concludes the proof for Proposition \ref{prop:emp}.
\end{proof} 

\subsection{Discussion on MC error} \label{appx:mcerror}
% % \begin{proof}
As in \citep{gelman1995bayesian}, the Monte Carlo approximation will be estimated to an error of approximately $\sqrt{\frac{Var\Big(\frac{1}{n_{k,e}}\sum_{x \in D_{k,e}}\mathbb{I}(h_{\theta^{T}_j}(x) > 0)\Big)}{N}}$, where $\mathbb{I}$ is the indicator function and we also have:
\begin{align}
    Var\Big(\frac{1}{n_{k,e}}\sum_{x \in D_{k,e}}\mathbb{I}(h_{\theta^{T}_j}(x) > 0)\Big) = \frac{1}{n_{k,e}^2}\sum_{x \in D_{k,e}}Var \big(\mathbb{I}(h_{\theta^{T}_j}(x) > 0)\big)
\end{align}
with $\mathbb{I}(h_{\theta^{T}_j}(x) > 0)$ is a random variable distributed by Bernoulli distribution $Bern\Big(Pr(z > 0 | \xi)\Big)$.

Therefore, $Var(\mathbb{I}(h_{\theta^{T}_j}(x) > 0)) = 
Pr(z > 0 | \xi)[1 - Pr(z > 0 | \xi)] \le 1/4$. As a result, the Monte Carlo approximation error is approximately $\frac{1}{2n_{k,e}\sqrt{N}}$.
% \end{proof}
% Note that the true data distribution of different groups and the DP-preserving noise distribution is intractable at inference time in real-world applications. Therefore, we provide a fairness certification under a finite number of data samples $D_{k,e}, k \in [K]$ associated with a random event $e$ at inference time in Appx. \ref{appx:cert_finite} with a marginal error. This sheds light on understanding our fairness certification better.

%\hai{These points do not sound interesting. You may want to link them with the DP-preserving noise scale to achieve desirable numbers of updating steps.}

% \noindent\framebox{
% \begin{minipage}{.96\linewidth}
% \textbf{Remark 2.} 
%     Practitioners can leverage our results to better balance the trade-offs among privacy, fairness, and utility by adaptively adjusting the training process of \textsc{FairDP}. For example, applying optimizers like Adam \citep{kingma2014adam} at the onset of training may enhance model utility and convergence rate under the same DP protection. As the model nears convergence, practitioners can transition to SGD to secure fairness certification, enabling us to overcome tight constraints on the weights of the last layer. Also, practitioners can adjust the hyper-parameter $M$ to achieve better fairness, such that the smaller $M$, the fairer the model is. However, small $M$ could degrade model utility since it constrains the decision boundary in smaller parameter space (see Table \ref{tab:impact-M} for details).
% \end{minipage}}

% \newpage
\section{Supplemental Results}
\label{Appx:supp}

\begin{table}[h]
    \caption{Impact of hyper-parameter $M$ on \textsc{FairDP} under $\epsilon =1.0$.}
    \label{tab:impact-M}
    \begin{center}
    \resizebox{0.6\columnwidth}{!}{%
    \begin{tabular}{lcccccc}
    \toprule
    Dataset & $M$ & AUC & Acc & DP & EOpp & EOdd \\
    \midrule
    \multirow{4}{*}{Adult} & 0.25 & $0.59$ & $0.75$ & $0.0$ & $0.0$ & $0.0$ \\
    & 0.50 & $0.77$ & $0.76$ & $0.001$ & $0.002$ & $0.001$ \\
    & 0.75 & $0.79$ & $0.79$ & $0.019$ & $0.021$ & $0.005$ \\
    & 1.00 & $0.80$ & $0.80$ & $0.027$ & $0.038$ & $0.011$ \\
    \midrule
    \multirow{4}{*}{Default-CCC} & 0.25 & $0.73$ & $0.78$ & $0.0$ & $0.0$ & $0.0$ \\
    & 0.50 & $0.75$ & $0.80$ & $0.011$ & $0.026$ & $0.009$ \\
    & 0.75 & $0.76$ & $0.81$ & $0.013$ & $0.018$ & $0.009$ \\
    & 1.00 & $0.76$ & $0.81$ & $0.016$ & $0.016$ & $0.009$ \\
    \midrule
    \multirow{4}{*}{UTK-Face} & 0.25 & $0.73$ & $0.64$ & $0.0$ & $0.0$ & $0.0$ \\
    & 0.50 & $0.79$ & $0.71$ & $0.087$ & $0.090$ & $0.044$ \\
    & 0.75 & $0.80$ & $0.73$ & $0.115$ & $0.104$ & $0.061$ \\
    & 1.00 & $0.80$ & $0.75$ & $0.130$ & $0.115$ & $0.068$\\
    \bottomrule
    \end{tabular}
    }
    \end{center} % \vspace{-10pt}
\end{table}

\end{document}